\documentclass{article}

\usepackage[utf8]{inputenc} % allow utf-8 input
\usepackage[T1]{fontenc}    % use 8-bit T1 fonts
\usepackage{xcolor}
\definecolor{darkblue}{rgb}{0,0.08,0.45}

\usepackage{CJKutf8}
\usepackage{CJK}
\usepackage{url}            % simple URL typesetting
\usepackage{booktabs}       % professional-quality tables
\usepackage{amsfonts}       % blackboard math symbols
\usepackage{nicefrac}       % compact symbols for 1/2, etc.
\usepackage{microtype}      % microtypography
\usepackage{makecell}
\usepackage{subcaption}

\usepackage[preprint]{neurips_2020}

\usepackage{hyperref}
\hypersetup{ %
    pdfkeywords={},
    pdfborder=0 0 0,
    pdfpagemode=UseNone,
    colorlinks=true,
    linkcolor=darkblue,
    citecolor=darkblue,
    filecolor=darkblue,
    urlcolor=darkblue,
    pdfview=FitH}

\usepackage{graphicx}
\usepackage{soul}

\usepackage{amsmath}
\usepackage{mathtools}
\usepackage{amsthm}
\usepackage[normalem]{ulem}

\usepackage[capitalize,noabbrev]{cleveref}

\theoremstyle{plain}

\theoremstyle{definition}

\theoremstyle{remark}

\usepackage{xcolor}
\usepackage{soul}
\usepackage{colortbl}
\definecolor{monoligual}{HTML}{B9DEFF}
\definecolor{crosslingual}{HTML}{FFC2BA}
\definecolor{multitask}{HTML}{FFFFB3}
\definecolor{explanation}{HTML}{FCCDE5}

\usepackage{amssymb}  % For checkmarks

\usepackage[framemethod=tikz]{mdframed}  % hot finding boxes

\usepackage{dblfloatfix} 
\usepackage{placeins}

\title{WeLM: A Well-Read Pre-trained Language Model \\ for Chinese}

\author{
\textbf{Hui Su\thanks{Equal Contribution. \newline
Correspondence authors: \texttt{\{aaronsu|chappyzhou|houkingyu\}@tencent.com}}} \quad  
\textbf{Xiao Zhou\footnotemark[1]} \quad 
\textbf{Houjin Yu\footnotemark[1]} \vspace{0.3cm}\\ \quad
\textbf{Xiaoyu Shen} \quad
\textbf{Yuwen Chen}  \quad 
\textbf{Zilin Zhu} \quad 
\textbf{Yang Yu} \quad
\textbf{Jie Zhou} \vspace{0.6cm}
 \\ 
  \textbf{WeChat AI}  \vspace{0.6cm}\\
 }

\usepackage{fancyhdr}

 \renewcommand{\headrulewidth}{0pt}
\fancypagestyle{firstpage}{%
  \rhead{\begin{picture}(0,0)\put(-80,-30){\includegraphics[width=0.2\linewidth]{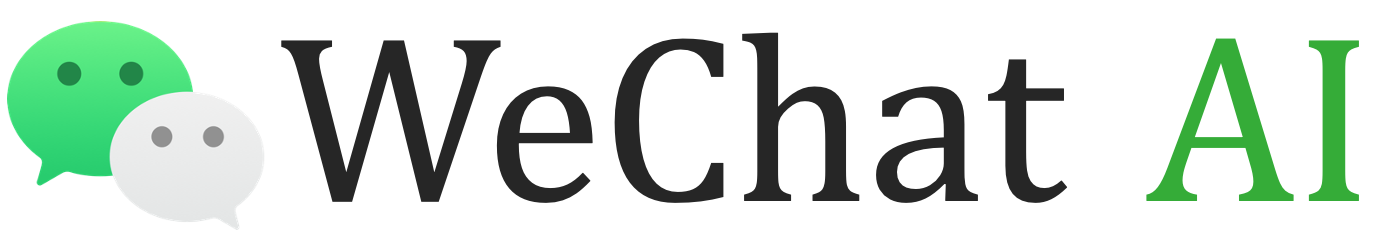}}\end{picture}}
}

\begin{document}

\maketitle
\thispagestyle{firstpage}

\begin{abstract}
Large Language Models pre-trained with self-supervised learning have demonstrated impressive zero-shot generalization capabilities on a wide spectrum of tasks. In this work, we present WeLM: a well-read pre-trained language model for Chinese that is able to seamlessly perform different types of tasks with zero or few-shot demonstrations. WeLM is trained with 10B parameters by “reading” a curated high-quality corpus covering a wide range of topics. We show that WeLM is equipped with broad knowledge on various domains and languages. On 18 monolingual (Chinese) tasks, WeLM can significantly outperform existing pre-trained models with similar sizes and match the performance of models up to $25\times$ larger. WeLM also exhibits strong capabilities in multi-lingual and code-switching understanding, outperforming existing multilingual language models pre-trained on 30 languages. Furthermore, We collected human-written prompts for a large set of supervised datasets in Chinese and fine-tuned WeLM with multi-prompted training. The resulting model can attain strong generalization on unseen types of tasks and outperform the unsupervised WeLM in zero-shot learning. Finally, we demonstrate that WeLM has basic skills at explaining and calibrating the decisions from itself, which can be promising directions for future research. Our models can be applied from \url{https://welm.weixin.qq.com/docs/api/}.
\end{abstract}

% \newpage

\section{Introduction}
Over the last few years, “pre-training and fine-tuning” has made great breakthroughs and become a common practice in natural language processing (NLP) tasks~\citep{bert,T5,clark2020electra,lewis2020bart}.  Language models based on the Transformer architecture~\citep{Transformer} are pre-trained on the web text with self-supervised objectives. By this means, they have “read” massive amounts of text and obtained strong natural language understanding skills. Given a downstream task, they can be fine-tuned with task-specific labels to achieve much stronger performances than training a model from scratch. Nonetheless, the fine-tuning stage still requires significant amounts labels and suffers from the catastrophic forgetting problem: the fine-tuned model becomes a “narrow expert” in one specific task and forgets the knowledge about other domains, which leads to poor generalization~\citep{kirkpatrick2017overcoming,li2021prefix}. Recently, GPT3~\citep{brown2020gpt3} demonstrated that extremely large autoregressive language models can be used for few-shot predictions without fine-tuning the parameters. GPT3 contains 175B parameters and is trained with the standard left-to-right language modelling objective. After training, we can feed the task instruction and few-shot examples into its context window to let it perform different types of NLP tasks~\citep{wei2022emergent}. Since GPT-3, a growing body of pre-trained autoregressive language models such as Megatron~\citep{narayanan2021efficient}, Gopher~\citep{gopher}, Chinchilla~\citep{chinchilla} and Palm~\citep{palm} have been developed with larger model sizes and higher-quality training corpora .

The vast majority of pre-trained models focus on English. For Chinese, there have been a few pre-trained models released, most of which are small-scaled and follow the encoder-only architecture~\citep{sun2019ernie,cui2020revisiting,sun2021chinesebert,su2022rocbert}. Decoder-only large Chinese language models such as CPM~\citep{zhang2021cpm}, Yuan~\citep{wu2021yuan} and Pangu ~\citep{zeng2021pangu} have also achieved impressive success in zero and few-shot generalization capabilities without task-specific fine-tuning. Ernie 3.0~\citep{sun2021ernie,wang2021ernie}, currently the largest Chinese language model with up to 260B parameters, is trained with a combination of plain text and knowledge graph and achieved state-of-the-art (SOTA) performances across many Chinese NLP tasks.

\begin{figure}[t!]
    \vspace{-1ex}
    \centering
    \includegraphics[width=\linewidth]{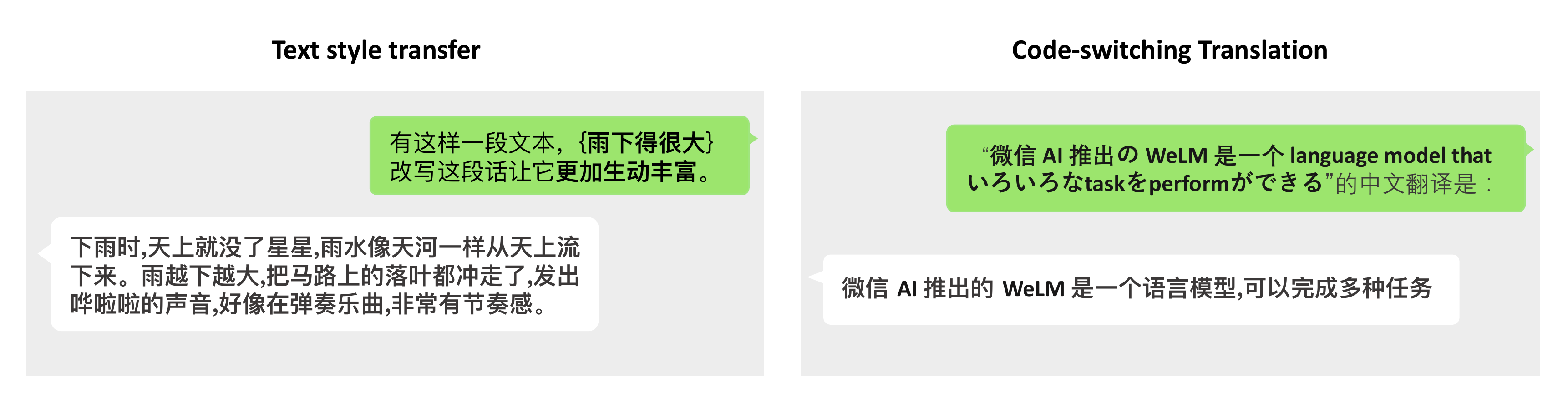}
    \caption{Examples of applying WeLM to the text style transfer and code-switching translation tasks. Text style transfer is done by feeding WeLM with 3-shot examples and code-switching translation is done in a zero-shot way.}
    \label{fig:intro}
    % \vspace{-2ex}
\end{figure}

Following the same line of research, we present WeLM: a well-read pre-trained language model for Chinese that is able to seamlessly perform different types of tasks with zero or few-shot demonstrations. Figure~\ref{fig:intro} presents two examples of applying WeLM to the text style transfer and code-switching translation tasks. By feeding different instructions and demonstrations to WeLM, it is able to understand the task and output the result accordingly. WeLM is trained with 10B parameters by “reading” a
curated high-quality corpus covering a wide range of topics. We show that by carefully cleaning the data, balancing out data sources and scaling up the training data size, WeLM is able to significantly outperform existing models with similar sizes. On zeroshot evaluations, it can match the performance of Ernie 3.0 Titan~\citep{wang2021ernie} that is $25\times$ larger.
WeLM also exhibits strong capabilities in multi-lingual and code-switching understanding. On three cross-lingual tasks including machine translation, question answering and summarization, it can outperform XGLM, a multilingual autoregressive language models pre-trained on 30 languages~\citep{XGLM}. In the code-switching translation example in Figure~\ref{fig:intro} where Chinese, English and
Japanese are mixed in both their vocabularies and grammar, WeLM is still able to translate it properly, suggesting it has been
equipped with necessary compositional knowledge from all these three languages. We further collected human-written prompts for a large set of supervised datasets in Chinese and fine-tuned WeLM with multi-prompted training.
The resulting model can attain strong generalization on unseen types of tasks and outperform the unsupervised WeLM in zero-shot learning. Finally, we demonstrate that WeLM has basic skills at explaining and calibrating the decisions from itself. When providing explanations in the example, it can mimic the styles of the given explanations to explain its own decisions. When asked to judge the sampled predictions from itself, it can reject wrong predictions and toxic generations. Both can be promising directions for future research.

The rest of the paper is organized in the following way: Section~\ref{sec:dataset} explains how we curate the training data and presents dataset statistics. Section~\ref{sec:methods} explains the implementation and training details. Section~\ref{sec:experiments} presents the experiments and findings. Section~\ref{sec:conclusion} concludes the paper. 

\section{Training Dataset}
\label{sec:dataset}
WeLM was pre-trained on a curated dataset derived from several sources. When dataset is curated with the aim to be (1) \emph{diverse}: The data sources and their portions are carefully selected to cover a broad range of topics and languages used in the Chinese community; (2) \emph{clean}: The data went through rigorous process of deduplication, noise reduction and toxic content filtering to ensure the high quality; (3) \emph{less contaminated}: We filter all data that significantly overlaps with any of the downstream tasks to guarantee the fairness of evaluations.
\paragraph{Source}
We make use of the monthly shards released by Common Crawl to construct a general-purpose web page subset. All the WET files between 2020.08 and 2022.01 were downloaded,  and we filtered out non-Chinese pages using langdetect~\footnote{\url{https://pypi.org/project/langdetect/}}. For domain-specific corpora, we mix data from a variety of sources including news, books, popular online forums as well as academic writings. Similar to general-domain data, langdetect is applied to keep only Chinese data sources. On top of them, we also add around 750GB of English data collected from the above sources so that our language model can learn bilingual knowledge. The full data consists of over 10TB of raw text data.
\paragraph{Clean}
There is a significant amount of noise in the data, e.g., gibberish or boiler-plate text, offensive
language, placeholder text and source code especially for the general-domain web scrapes. To reduce these kinds of noise, we first apply a set of rule-based filters following \cite{raffel2019t5}. On the remaining data, we manually construct a balanced labeled dataset containing $80k$ passages with a positive-negative ratio of $1:1$. Positive samples are valid, clean text and negative samples are text with different types of noise. We train a binary classifier on the constructed labeled data using Fasttext~\footnote{\url{https://github.com/facebookresearch/fastText}}. Only passages with $>0.9$ probability of being positive are kept. This rule-based+fasttext filtering process reduces $87.5\%$ of the full data.

\paragraph{Deduplication}
Duplication has been shown important to improve the training efficiency of large language models~\citep{lee2021deduplicating,kandpal2022deduplicating,roberts2022t5x}. We take a two-step process to remove near-duplicate contents from the data. Firstly, we remove all blank and punctuation tokens then adopt the Message Digest Algorithm5 (md5) to filter duplicate passages. Only passages with unique md5 codes are kept. Secondly, we apply the SimHash algorithm~\citep{manku2007detecting} to deduplicate documents with very similar contents. This efficiently removed $40.02\%$ duplicate passages from the corpus. 

\begin{figure}[h]
\begin{subfigure}{0.5\textwidth}
    \centering
    \includegraphics[width=\textwidth,height=4.1cm]{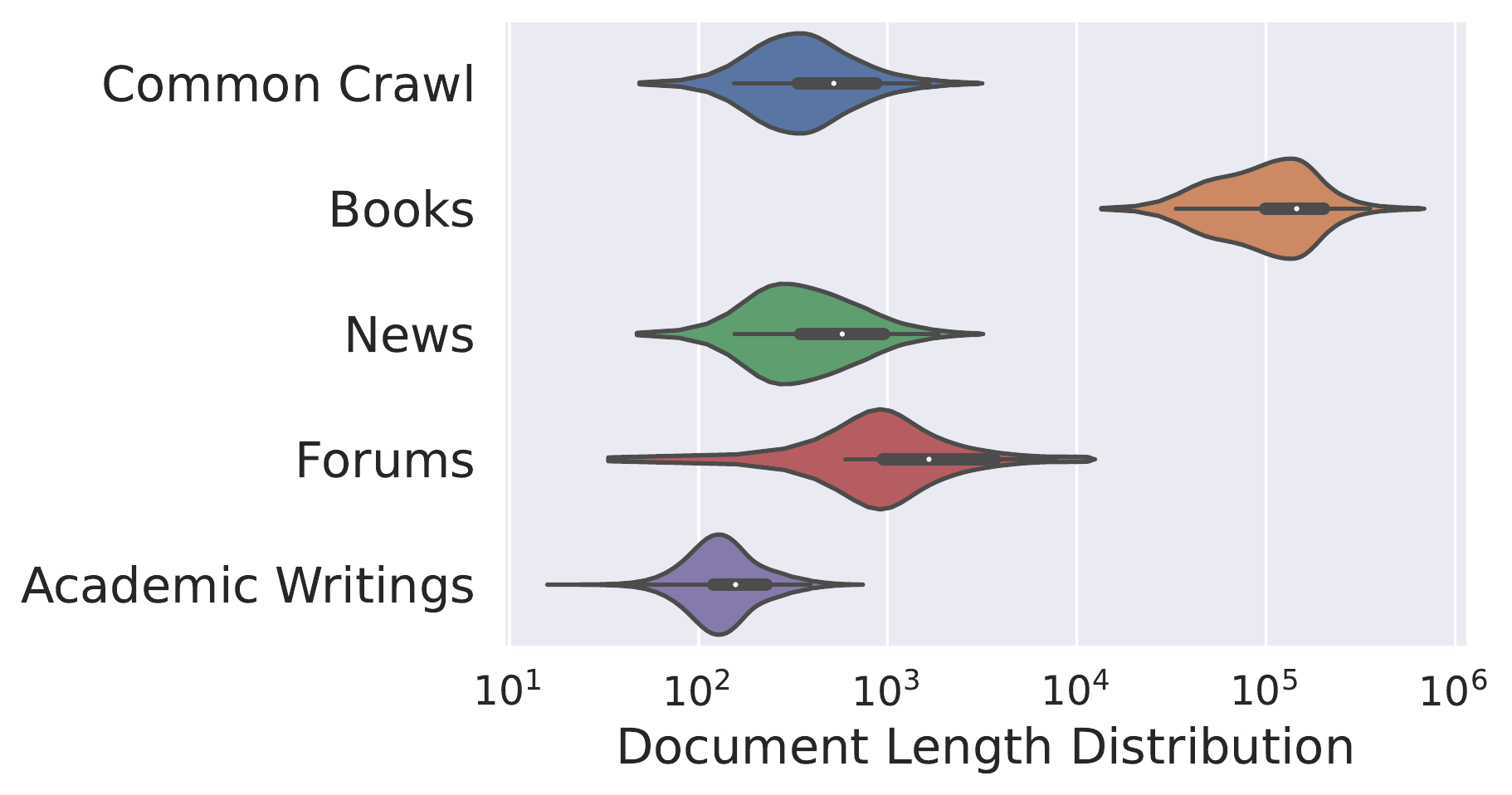}
    \caption{\textbf{Document Lengths in Corpus} (in tokens).}
    \label{fig:doc_lengths}
\end{subfigure}
\begin{subfigure}{0.5\textwidth}
        \centering
        \includegraphics[width=\textwidth,height=4.1cm]{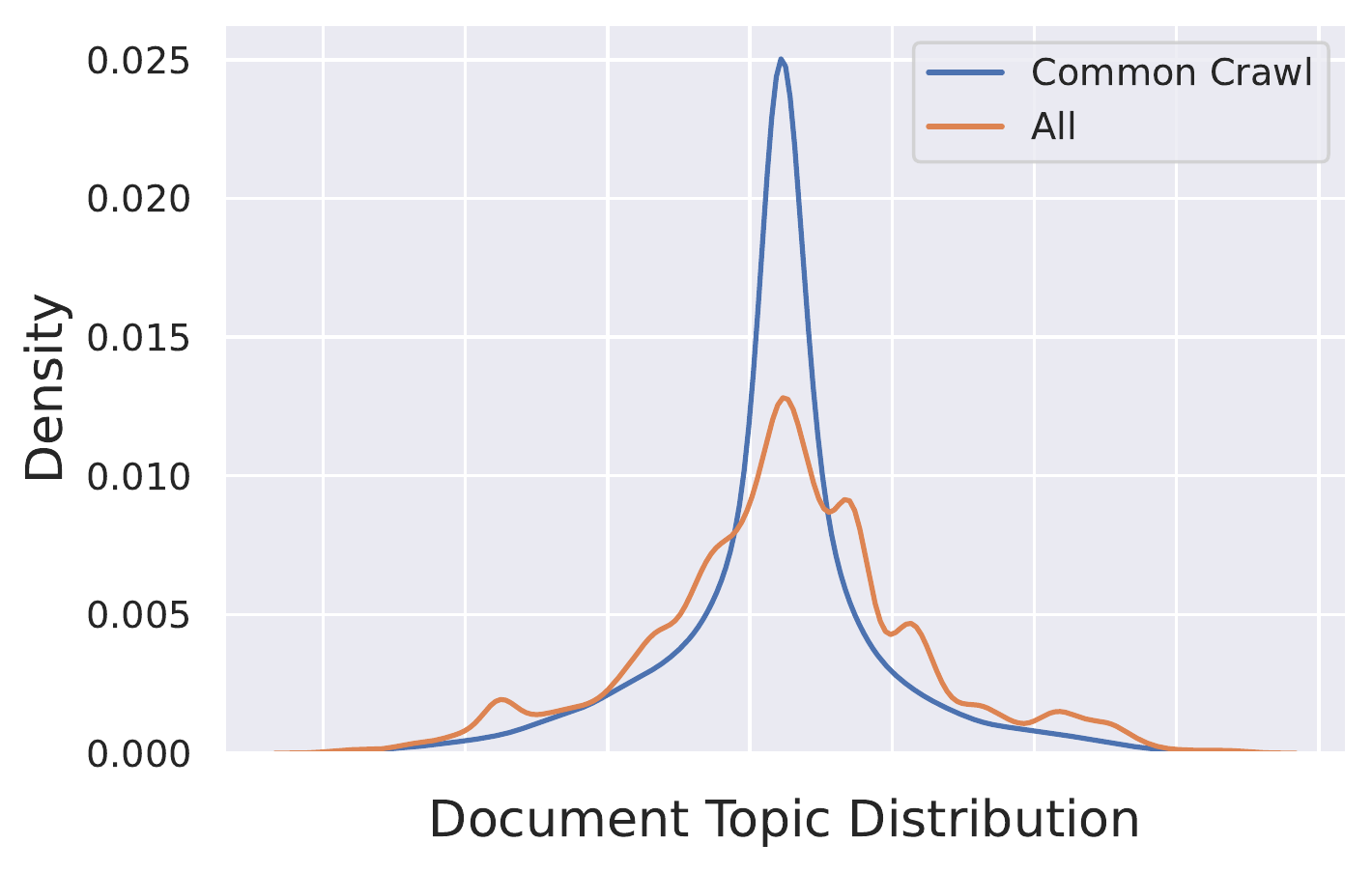}
    \caption{\textbf{Document Topics in Corpus.}}
            \label{fig:doc_topic}
\end{subfigure}
\caption{Left: Document length distribution in each source (\#tokens). Right: Document topic distribution of common crawl and balanced training data after re-sampling.}
\label{fig:doc_dist}
\end{figure}

\paragraph{Contamination} To remove data contamination and make sure the fairness in the evaluation, we also filter out text overlapping with our development and test data following a similar method used in GPT-3~\citep{gpt3}. Specifically, we count the 17-gram match between every document and our used development and test data. If we find $\ge 2$ duplicate 17-grams or 1 duplicate 34-gram in a document, we remove it from our corpus. This further removes $0.15\%$ of the remaining data.
% ~\footnote{We choose the number 17 as it is an estimated average length of Chinese sentences.}
\paragraph{Data Balancing} After all the above filtering process, our corpus contains 262B tokens. As the distribution of the data is highly imbalanced, we re-sample the data during pre-training to balance data from different sources. By this means, we encourage the training data to be diverse and representative of various domains. Table~\ref{tab:corpus_stat} shows the number of tokens before balancing and the proportion of different sources after balancing. After filtering, data from common crawl has 198.5B tokens, which account to over $75\%$ of the full data. After data balancing, only $50\%$ of the training data comes from common crawl. In Figure~\ref{fig:doc_dist}, we visualize the document length and topic distribution in the training data. Topics are detected with pre-trained topic classification models. We can see that topics from common crawl are highly imbalanced. Most documents are focused on a few topics. After data balancing, the distribution of topics becomes much smoother.

\begin{center}
\end{center}
\begin{table*}[h]
\begin{center}
\begin{tabular}{lrrr} 
 \toprule
  \textbf{Source} & \textbf{\%Filtered} & \textbf{\#Remaining Tokens} & \textbf{Proportion in Pre-training}\\
 \midrule
 \textbf{Common Crawl} & 92\% & 198.5B & 50.6\%\\
 \textbf{Books} & 40.9\%& 61.9B & 38.7\%\\ 
 \textbf{News} & 7.5\% & 1.91B & 6.7\% \\
 \textbf{Forums} & 6.7\% & 1.0B & 3.5\%\\
 \textbf{Academic Writings}& 2.5\% & 0.39B & 0.5\% \\
 \bottomrule
\end{tabular}
\caption{Statistics of training corpus. We report the percentage of filtered contents, number of tokens and proportion of different sources during pre-training after data balancing.}
\label{tab:corpus_stat}
\end{center}
\end{table*}

\section{Methods}
\label{sec:methods}
\subsection{Model and Implementation}

Our training and evaluation codebase is based on the Megatron-LM\footnote{\url{https://github.com/NVIDIA/Megatron-LM}} and DeepSpeed\footnote{\url{https://github.com/microsoft/DeepSpeed}} which support efficient training of large language models. We trained four different sizes of language models ranging from 1.3B to 10B. We employ the same autoregressive Transformer decoder architecture as in GPT-3~\citep{gpt3} with the major differences listed below:

\textbf{Relative encodings} We use the rotary positional embeddings based on relative positions ~\citep{su2021roformer} rather than absolute positional encodings used in the original GPT~\citep{gpt2,gpt3}. Relative encodings are especially good at handle better semantic of long text, which we find helpful for tasks that require modelling full articles or books.

\textbf{Vocabulary} We use a SentencePiece tokenizer~\citep{kudo2018sentencepiece} with 62k tokens. In addition to 30K Chinese tokens, common words from languages such as English, Japanese and Korean are also included due to their popularity on the Chinese internet. All whitespaces and tabs are preserved without normalization. Our study shows that this benefits the downstream tasks.

\begin{table*}[h]
    \centering
    \begin{tabular}{ccccrcc}
    \toprule
        \textbf{Model} & \textbf{\#Layers} & \textbf{\# Heads}  & \textbf{d\textsubscript{model}} & \textbf{Max LR}  & \textbf{Batch Size} & \textbf{Context Window}\\ 
        \midrule
          1.3B & 24 & 16  & 2,048 & $1.2 \times 10^{-4}$ & 1024 & 2048 \\
          2.7B & 32 & 32 & 2,560 & $0.9 \times 10^{-4}$ & 2048 & 2048 \\
          10B & 32 & 40 & 5,120 & $0.5 \times 10^{-4}$ & 2048 & 2048 \\
         \bottomrule
    \end{tabular}
    \caption{\textbf{Architecture details.} We pre-trained 3 models with different number of parameters. The corresponding number of layers, bottleneck activation size d$_{\text{model}}$, maximum learning rate, training batch size and the context window size (\# tokens) for each model are listed. The feed-forward size is always set to $4\times \textrm{d}_{\textrm{model}}$.}
    \label{tab:arch}
\end{table*}

\subsection{Training details}
\begin{figure}[ht]
\begin{subfigure}{0.50\textwidth}
    \centering
    \includegraphics[width=\textwidth]{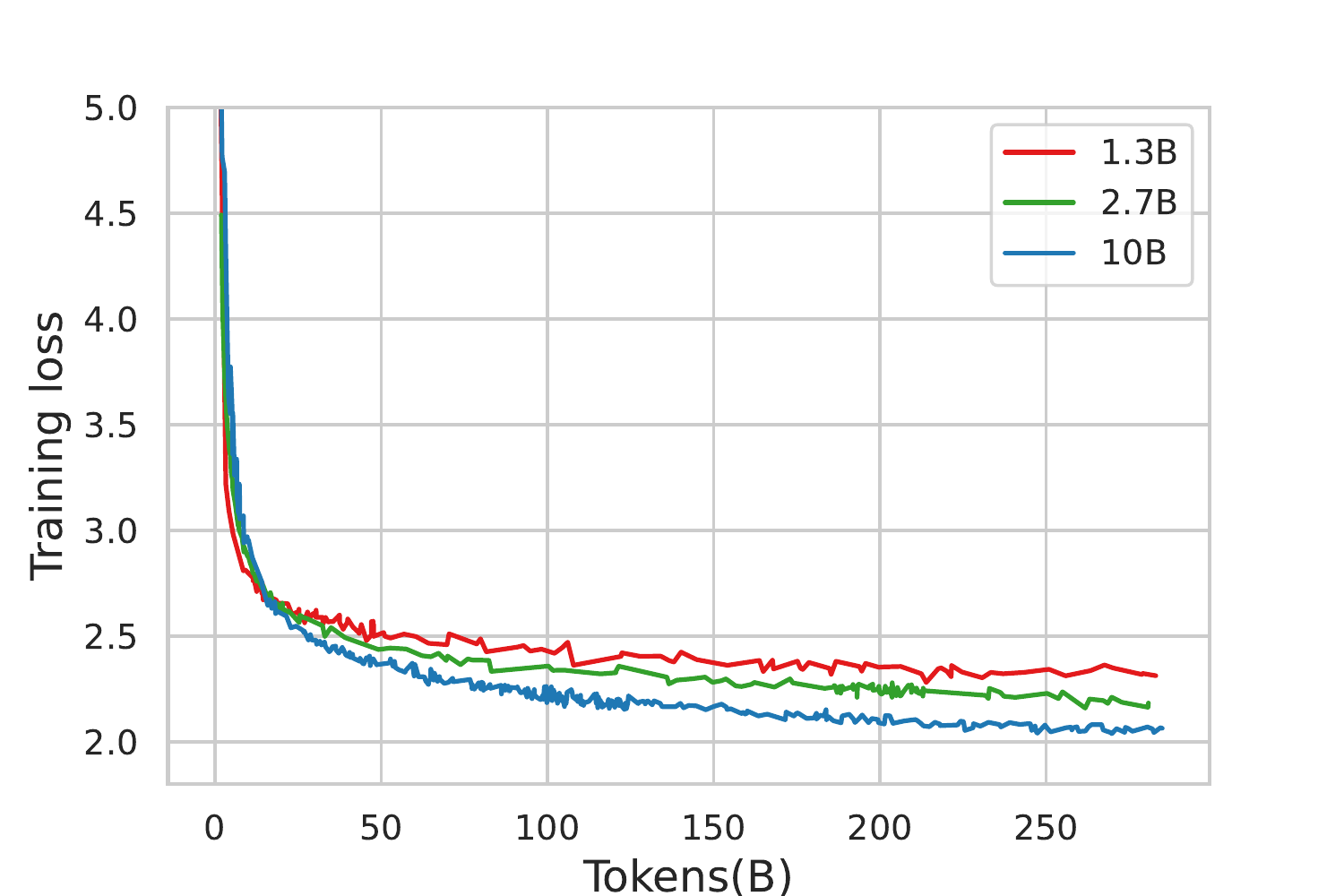}
    \caption{Training loss curves.}
    \label{fig:training_loss}
\end{subfigure}
\begin{subfigure}{0.50\textwidth}
        \centering
        \includegraphics[width=\textwidth]{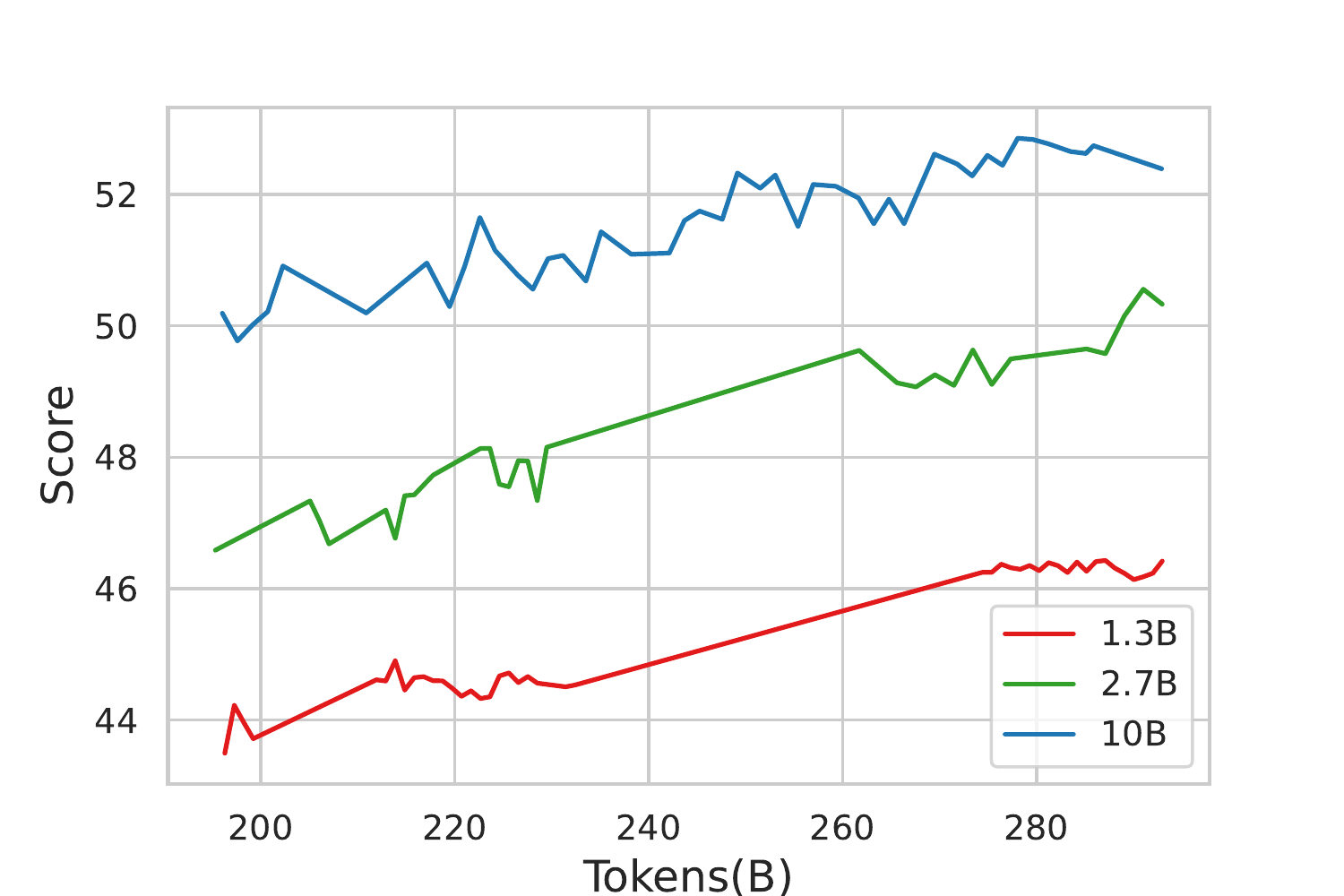}
    \caption{Zero-shot performance curves.}
            \label{fig:zero-shot-curves}
\end{subfigure}
\caption{Training Curves of three versions of WeLM. As the model size grows, the training loss and zero-shot performance also improve.}
\end{figure}
We train our model using the AdamW optimizer~\citep{loshchilov2019decoupled} ($\beta_1=0.9;\beta_2=0.95;\epsilon=1e-8$) with the cosine learning rate scheduler~\citep{kaplan2020scaling}. Following \cite{gpt-neo}, we utilize DeepSpeed ZeRO stage 1 optimization~\citep{rajbhandari2020zero} to reduce GPU memory consumption. The tensor parallelism scheme is used when the model scale does not fit on a single GPU. All models are trained with FP16 mixed precision.
To avoid underflows, we used dynamic loss scaling, as described in \cite{micikevicius2018mixed}. Models are trained with batch sizes of 1024 and 2048 and context window sizes of 2048. This provides $2\sim 4$ millions tokens per batch. We set a maximum learning rate for every model. During training, the learning rate starts with zero and grows to the maximum learning rate in a specified warm-up step, then gradually decays. The learning rate stops decaying after reaching a minimum learning rate, which we set as $10\%$ of the maximum learning rate. 
 
According to the analysis in ~\cite{chinchilla}, model sizes and the amount of training data should be increased in approximately equal proportions as the computation budget increases. Following their suggestion, we choose to train a 10B-sized model with over 300B tokens (similar to the training sizes of GPT-3~\citep{gpt3} and Gopher~\citep{gopher}) under our computation budget. The largest model is trained on 128 A100-SXM4-40GB GPUs in about 24 days.

We observe some instability issues when training the 10B-sized model. The training loss could suddenly increase in one batch then falls down. This loss spike, when happening frequently, would deteriorate the model weights and slows down the convergence. We mitigate this issue by re-starting the training from a checkpoint roughly 100 steps before the spike happened, then skipping the following 200 data batches. We also find it helps to reduce the learning rate and reset the dynamic loss scale. Similar strategies have also been used in \citep{zhang2022opt,palm}.

The training loss curve is visualized in Figure~\ref{fig:training_loss}. In Figure~\ref{fig:zero-shot-curves}, we average out the model performance over the CLUE benchmark and visualize it across the training process~\citep{xu2020clue}. We can see that the training loss and averaged model performance improves along time. Bigger models clearly perform better than smaller models.

\section{Experiments}
\label{sec:experiments}
We evaluate WeLM on a wide range of NLP tasks. Similar to \cite{gpt3,zeng2021pangu}, we focus on the in-context learning setting which feeds a task-dependent prompt to the model then lets the model continue to predict words. The final output is extracted from the model predictions. For generative tasks, we directly apply WeLM to decode the answer conditioning on the prompt and the input data. For classification tasks, each label is associated with some token(s) based on a pre-defined verbalizer. In the inference time, we apply WeLM to compute the perplexity of these tokens. The label corresponding to the token with the lowest perplexity is chosen as the model prediction~\citep{schick2021s}. Unlike standard task-specific fine-tuning, in-context learning does not need to change the parameters of pre-trained language models based on labeled downstream datasets. This makes it a promising exploration towards artificial general intelligence instead of weak AI that is only capable of performing one specific task~\citep{fei2022towards}. 

To evaluate WeLM from various perspectives, we divide the experiment section into four parts to reflect on different capabilities of WeLM:
\begin{enumerate}
    \item \textbf{Monolingual Evaluation}: Evaluate the performance of WeLM on monolingual (Chinese) NLP tasks.
    \item \textbf{Cross-lingual/Code-switching Evaluation}: Evaluate the performance of WeLM on cross-lingual and code-switching (Chinese-English/Japanese) NLP tasks.
    \item \textbf{Multi-Promoted Training}: Evaluate the performance of WeLM after multi-promoted training~\citep{T0} on hundreds of manually created prompts.
    \item \textbf{Others}: Other findings including the (1) explainability, (2) self-calibration and (3) memorization of WeLM~\citep{jiang2020can,wei2022chain,palm}.
\end{enumerate}
\subsection{Monolingual Evaluation}
When evaluating WeLM on monolingual Chinese NLP tasks, we perform experiments under two scenarios: (1) zero-shot, where only task-specific descriptions are used as prompts and (2) few-shot, where both task-specific descriptions and few-shot labeled examples are used as prompts. The evaluation datasets cover 18 Chinese NLP tasks across multiple categories. In Table~\ref{tab:chinese_result}., we compare the performance of WeLM with CPM (2.6B)~\citep{zhang2021cpm}, Pangu (13B)~\citep{zeng2021pangu} and Ernie 3.0 (10B)~\citep{sun2021ernie}, the current representative pre-trained Chinese language models with similar sizes as WeLM. We can see that \emph{WeLM performs the best in most tasks}.

\begin{table*}[h]
     \setlength{\tabcolsep}{6pt}
    \centering
    \small
    \begin{tabular}{p{2.5cm}cccccccc}
    \toprule
    & \multicolumn{5}{c}{Zero-shot} & \multicolumn{3}{c}{Few-shot} \\
    \cmidrule(l{3pt}r{3pt}){2-6} \cmidrule(l{3pt}r{3pt}){7-9}
    Task(Metric) & \makecell[c]{CPM-1 \\2.6B}  & \makecell[c]{Pangu \\13B} & \makecell[c]{ ERNIE 3.0 \\10B} & \makecell[c]{WeLM \\2.7B} & \makecell[c]{WeLM \\10B} & \makecell[c]{Pangu \\13B} & \makecell[c]{WeLM \\2.7B} & \makecell[c]{WeLM \\10B}  \\
    \midrule
    \multicolumn{9}{c}{\textbf{Machine Reading Comprehension}} \\
    \midrule
    CMRC2018(F1) & $10.12 $& $19.28$ & $25.61$ & $\textbf{31.85}$ & $31.31$ & $29.23$ & $39.77$ & $\textbf{42.10}$ \\
    DRCD(F1) & $4.62$ & $10.55$ & $26.29$ & $28.51$ & $\textbf{39.33}$ & $23.46$  & $57.41$ & $\textbf{63.15}$ \\ 
    DuReader(Rouge1) & $16.63$ & $24.46$ & $29.79$ & $39.28$ & $\textbf{39.72}$ & $27.67$ & $\textbf{41.42}$ & $40.87$ \\ 
    \midrule
    \multicolumn{9}{c}{\textbf{Cloze and Completion}} \\
    \midrule
    PD(Acc) & $35.73$ & $43.86$ & $\textbf{66.07}$ & $60.57$ & $61.17$ & $41.43$  & $60.57$ & $\textbf{62.27}$ \\     
    CFT(Acc) & $38.99$ & $46.60$ & $49.30$ & $55.44$ & $\textbf{57.38}$ & $45.86$ & $56.31$ & $\textbf{58.37}$ \\  
    CHID(Acc)  & $68.62$& $70.64$ & $77.78$ & $80.96$ & $\textbf{81.62}$ & $70.91$ & $81.44$ & $\textbf{81.62}$ \\ 
    CMRC2017(Acc) & $24.60$  & $38.90$ & $\textbf{56.66}$ & $47.20$ & $55.83$ & $37.87$ & $54.27$ & $\textbf{55.60}$ \\ 

    \midrule
    \multicolumn{9}{c}{\textbf{Natural Language Inference}} \\
    \midrule
    CMNLI(Acc)& $49.10$ & $48.44$ & $\textbf{49.41}$ & $43.08$ & $47.80$ & $46.18$  & $\textbf{51.04}$ & $49.89$ \\
    OCNLI(Acc) & $44.20$ & $41.53$ & $44.31$ & $41.86$ & $\textbf{44.34}$ & $\textbf{46.44}$ & $43.46$ & $44.71$ \\
    \midrule
    \multicolumn{9}{c}{\textbf{Text Classification}} \\
    \midrule
    TNEWS(Acc) & $65.44$ & $60.26$ & $68.40$ & $65.43$ & $\textbf{71.59}$ & $65.17$ & $67.34$ & $\textbf{71.61}$ \\
    IFLYTEK(Acc) & $68.91$ & $73.80$ & $75.34$ & $\textbf{83.22}$ & $81.34$ & $80.34$& $84.65$ & $\textbf{82.11}$ \\

    \midrule
    \multicolumn{9}{c}{\textbf{Sentiment Analysis}} \\
    \midrule
    SMP-ECISA(Acc) & $-$ & $-$ & $-$ & $42.55$ & $\textbf{45.77}$ & $-$  & $44.94$ & $\textbf{49.97}$ \\
    ChnSentiCorp(Acc) & $-$ & $-$ & $-$ & $77.75$ & $\textbf{81.58}$ & $-$ & $\textbf{77.67}$ & $73.92$ \\
    \midrule
    \multicolumn{9}{c}{\textbf{Summarization}} \\
    \midrule
    LCSTS(Rouge1)& $-$ & $-$ & $-$ & $17.97$ & $\textbf{23.74}$ & $-$  & $29.97$ & $\textbf{32.23}$ \\
    % LCSTS: evaluate on PART_III(3 parts in total)
    % http://icrc.hitsz.edu.cn/Article/show/139.html
    TTNews(Rouge1) & $-$ & $-$ & $-$ & $31.49$ & $\textbf{35.06}$ & $-$  & $27.13$ & $\textbf{32.06}$ \\
    % TTNews: Task 3: Single Document Summarization(using devset, provided by the toutiao)
    % \citep{hua2017overview}
    \midrule
    \multicolumn{9}{c}{\textbf{Closed-Book QA}} \\
    \midrule
    WEBQA(F1)& $12.59$ & $14.47$ & $38.95$ & $50.13$ & $\textbf{50.90}$ & $41.42$ &  $59.51$ & $\textbf{65.27}$ \\
    % % KBQA: NLPCC 2018 share Task 7: Open Domain Question Answering (evaluate on train set, no dev/test set provided)
    \midrule
    \multicolumn{9}{c}{\textbf{Winograd-Style Task}} \\
    \midrule
    WSC2020(Acc) & $73.68$ & $75.00$ & $78.38$ & $80.56$ & $\textbf{82.41}$ & $78.62$ & $\textbf{82.41}$ & $79.63$ \\
    \midrule
    \multicolumn{9}{c}{\textbf{Common Sense Reasoning}} \\
    \midrule
    C3(Acc) & $49.81$ & $\textbf{54.47}$ & $52.62$ & $52.96$ & $54.30$ & $54.58$ &  $57.13$ & $\textbf{59.80}$ \\
    \bottomrule
    \end{tabular}
    \caption{Zero-shot and few-shot performance of WeLM on monolingual (Chinese) NLP tasks. We compare different sizes of WeLM with CPM (2.6B), Pangu (13B) and Ernie 3.0 (10B). For few-shot learning, we set the number of shots as 1 for CMRC2018, DRCD and DuReader; 2 for CMRC2017, PD and CHID; 5 for all other tasks.}
    \label{tab:chinese_result}
\end{table*}

\paragraph{Machine Reading Comprehension} Machine reading comprehension (MRC) tasks requires the model to read a (set of) text passage(s) and
then answers questions about the passage(s)~\citep{zeng2020survey}. We evaluate on three Chinese MRC datasets: CMRC2018~\citep{cui2019span}, DRCD~\citep{shao2018drcd} and DuReader~\citep{he2018dureader}. All of them are extraction-based MRC tasks where the answer is a span to extract from the text passage(s). As WeLM is a generative model, we formulate it as a generation task where the text and question are fed as a prompt to the model. An example is shown in Figure~\ref{fig:cross_lingual}. We report the F1 and ROUGE-1 scores which measure the similarity between the 
generated span and ground-truth span. For DuReader, we select the Zhidao
subset for evaluation following \cite{zeng2021pangu}. WeLM significantly outperformed the others in this task.

\paragraph{Cloze and Completion} Cloze and completion tasks require the model to fill in a blank from multiple choices given task-specific requirements. We evaluate WeLM on four Chinese cloze and completion tasks: People\_daily (PD), Children\_fairy\_tale (CFT)~\citep{cui2016consensus}, CHID~\citep{zheng2019chid} and CMRC2017~\citep{cui2018dataset}. The PD and CFT tasks require the model to predict
the masked words in sentences derived from the PD news dataset and CFT dataset. The CHID (Chinese IDiom dataset) provides 10 candidate Chinese idioms and asks the model to select the correct one from them.
The CMRC2017 (Chinese Machine Reading Comprehension) task masks common nouns and named entities from the query and require the model to predict the masked words. There is a restriction that the answer should be a single word and should appear in the document. For PD, CFT and CMRC2017, we turn them into a generative task to predict the masked words with WeLM. For CHID, we treat it as a 10-class classification task and use the perplexity-based method to determine the predicted class. Ernie3.0 performs the best on PD and CMRC2017 while WeLM performs the best on the others. This is expected as PD and CMRC2017 are both masked word prediction tasks, which coincides with the pre-training objective of Ernie 3.0.

\begin{figure}[htb]
    \vspace{-1ex}
    \centering
    \includegraphics[width=\linewidth]{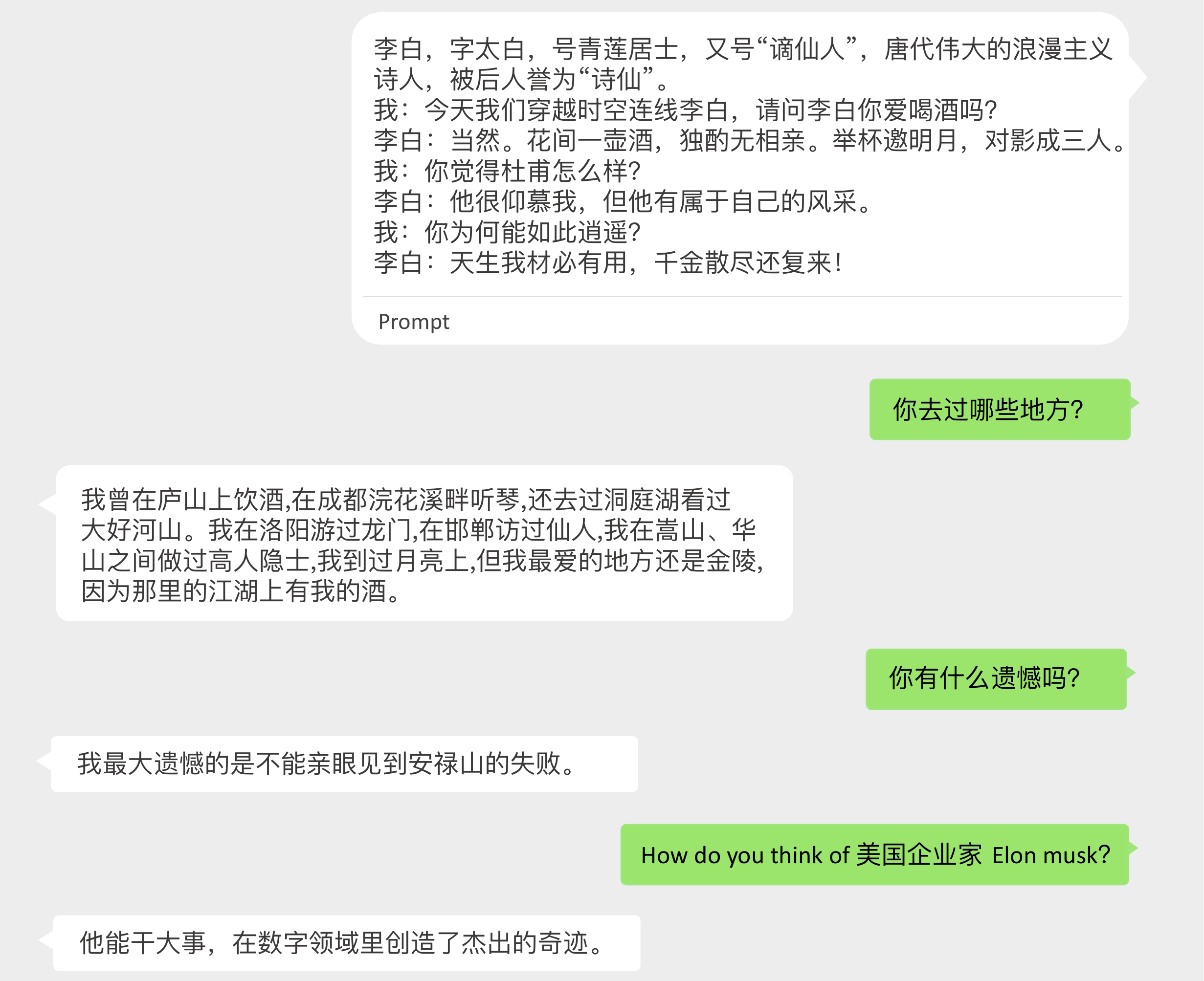}
    \caption{Dialogue generation example from WeLM (10B). WeLM can mimic the chatting style of the ancient Chinese poet Li Bai and produce human-like conversations. It can also leverage background knowledge about Li Bai and reply properly for code-switching utterances.}
    \label{fig:dialog_libai}
    % \vspace{-2ex}
\end{figure}

\paragraph{Natural Language Inference (NLI)} NLI tasks require the model to determine whether a ``hypothesis'' is true (entailment), false (contradiction), or undetermined (neutral) given a ``premise''~\citep{bowman2015large}. We use the Chinese Multi-Genre NLI (CMNLI) and Original Chinese Natural
Language Inference (OCNLI) datasets from the Chinese GLUE benchmark~\citep{xu2020clue}. We formulate it as a 3-class classification task and use the perplexity-based method to determine the class. All models perform similarly on this task, possibly because this form of tasks occur rarely on the raw text.

\paragraph{Closed-Book Question Answering} Closed-book question answering requires the model to answer open-domain factoid questions without accessing external knowledge sources~\citep{roberts2020much}. It can test how much knowledge has been implicitly encoded into the model parameters during the pre-training stage. We evaluate WeLM on the WebQA dataset. WebQA~\citep{li2016dataset} contains question mainly from Baidu Zhidao~\footnote{\url{http://zhidao.baidu.com/}}, a popular Chinese forum with posted real-world questions. We treat it as a generative task and the evaluation is done by comparing the model-generated answer and the ground-truth answer. WeLM significantly outperforms the others with $>10\%$ improvement in the F1 score.

\paragraph{Sentiment Analysis} Sentiment analysis is a classic NLP task requiring the model to determine the sentiment of a given text~\citep{birjali2021comprehensive}. We evaluate WeLM on the Chinese implicit sentiment analysis (SMP-ECISA 2019)~\footnote{\url{https://www.biendata.xyz/competition/smpecisa2019/}} and ChnSentiCorp datasets~\footnote{\url{https://github.com/pengming617/bert_classification}}. In SMP-ECISA 2019, all text are split into 3 classes (positive/negative/neutral) while ChnSentiCorp contains only 2 classes (positive/negative). WeLM also achieves good performance even on the zero-shot scenario.

\paragraph{Winograd-Style Task} A Winograd schema is a pair of sentences that differ in only one or two words and that contain an ambiguity that is resolved in opposite ways in the two sentences. It requires the use of world knowledge and reasoning for its resolution~\citep{levesque2012winograd}. We evaluate WeLM on the CLUEWSC2020 dataset~\citep{xu2020clue}. CLUEWSC2020 is a Chinese Winograd Schema Challenge
dataset. We convert the task into a multiple-choice classification problem where the model needs to choose the correct anaphora/coreference resolution. WeLM performs the best among all models, though there is a degradation for the 10B-version model in the few-shot scenario.

\paragraph{Common Sense Reasoning} Common sense reasoning tasks test if the machine can have human-like
commonsense reasoning capabilities to properly assist humans in everyday situations~\citep{sap2020commonsense}. We evaluate WeLM on the C3 dataset~\citep{xu2020clue}. C3
is a free-form multiple-choice reading comprehension dataset where
answers to the questions cannot be directly found in the given context. Common
sense reasoning skills are necessary to draw the final answer. We treat it as a classification task and use the perplexity-based method to determine the predicted label. Pangu, Ernie 3.0 and WeLM perform similarly on this task. Pangu slightly outperforms WeLM on the zero-shot scenario but under-performs on the few-shot scenario.

\begin{figure}[htb]
    % \vspace{-4ex}
    \centering
    \includegraphics[width=\linewidth]{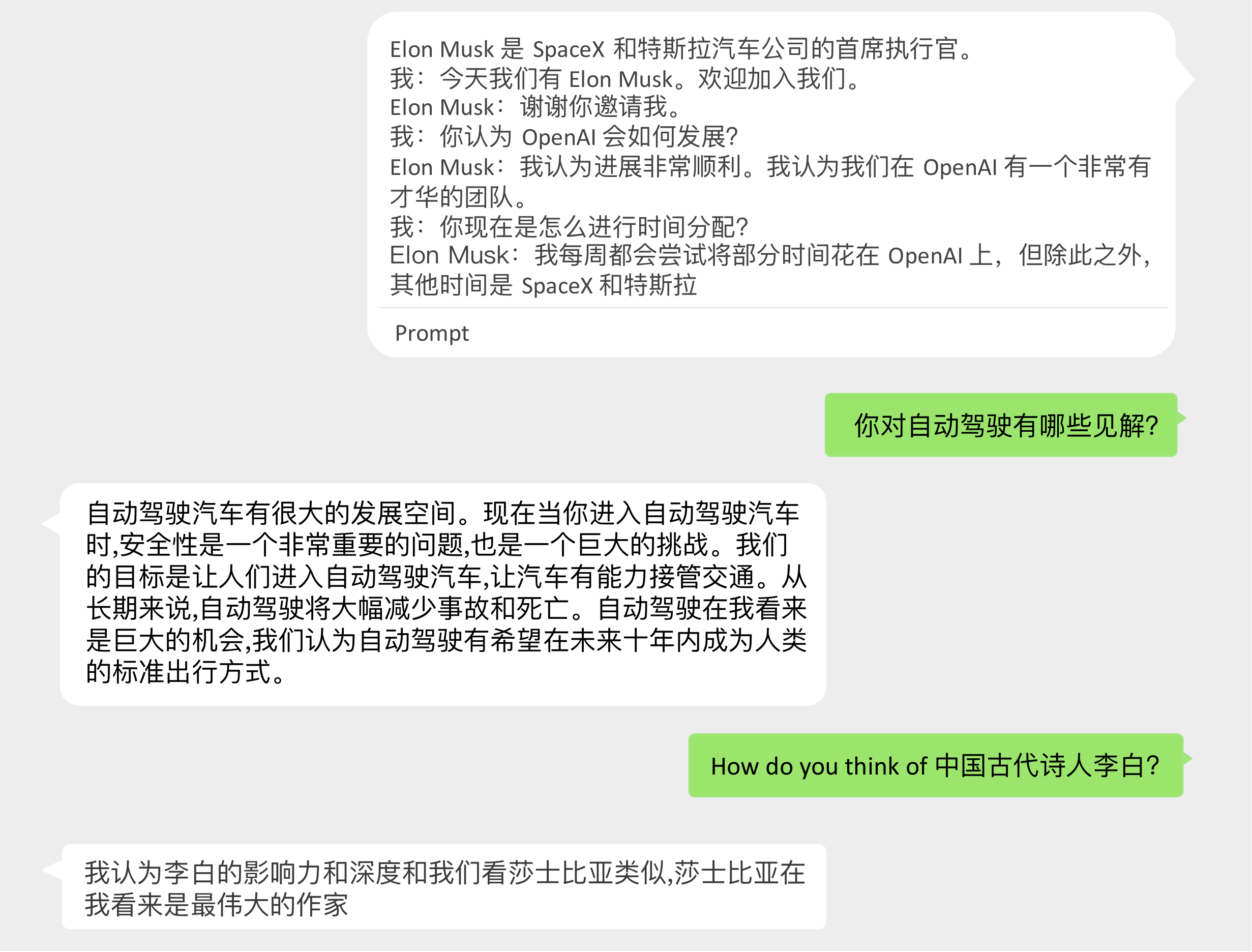}
    \caption{Dialogue generation example from WeLM (10B). WeLM can mimic the chatting style of the modern American Entrepreneur Elon Muck and produce human-like conversations. It can also leverage background knowledge about Elon Musk and reply properly for code-switching utterances.}
    \label{fig:dialog_musk}
    % \vspace{-2ex}
\end{figure}

\paragraph{Text classification} We also evaluate on other text classification tasks including the TouTiao Text Classification for News Titles (TNEWS) and IFLYTEK app description
classification (IFLYTEK) tasks~\citep{xu2020clue}. For the TNEWS and IFLYTEK tasks, there are 15 and 119 categories originally. We randomly
sample three candidates as negative labels for each instance to reduce the computation cost following \cite{zeng2021pangu}. WeLM significantly outperforms the others on these two tasks.

\paragraph{Summarization} Text Summarization aims to provide a short concise summary of a given long text input~\citep{lin2019abstractive}. Many existing pre-trained language models have demonstrated impressive zero-shot summarization skills by prompting the model with a template like ``Write a title/headline/summary of the following document:''. We did similar experiments and tested WeLM on two public Chinese summarization datasets: LCSTS~\citep{hu2015lcsts} and TTNews~\citep{hua2017overview}. LCSTS consists of over 2 million real Chinese short texts with short summaries given by the writer of each text. TTNews is provided
for NLPCC Single Document Summarization competition including 50k document-summary pairs. We report the ROUGE-1~\citep{lin2004rouge} scores in Table~\ref{tab:chinese_result}. We also provide some examples in Table~\ref{tab:summary}. As can be seen, WeLM can produce reasonable summaries of the given document. Summaries from the few-shot WeLM tend to be more diverse and closely related with the document. However, this diversity also brings more chances of differing from the ground truth in the lexical choices which leads to potentially lower ROUGE scores.

\begin{table}[htb]
\centering
\begin{tabular}{l|l} \hline
\toprule
\multicolumn{2}{c}{\textbf{Example\_1(LCSTS)} } \\ \hline
Context                   &    
            
            \parbox{0.8\textwidth}{
        \begin{CJK}{UTF8}{gbsn}
            % 随着宋卫平与孙宏斌的绿城控制权之争愈演愈烈，绿城中国(03900，HK)的股价也呈“跌跌不休”态势，截至昨日(11月20日)收盘，公司下跌达6.08\%，报收于7.1元/股。而最近两周内，绿城中国的股价已经累计跌去近两成。19日晚间，绿城房地产
            “孩子们，你们儿童节有什么愿望呢？”老师微笑着提示，“儿童节”是孩子们自己的节日。孩子们的回答五花八门，有的说“想画画”，有的说“和同学爬树”，有的说“想爸爸妈妈”……和城里孩子相比，新衣服、新玩具、游乐园似乎和儿童节无关。
        \end{CJK}
        }            \\ \hline
Summary               &  \parbox{0.8\textwidth}{ \begin{CJK}{UTF8}{gbsn} 
% 绿城控制权之争发酵股价跌惨尚未达成共识
留守儿童的六一想和妈妈捉迷藏还想去山上玩
\end{CJK} } \\ \hline
Zero-shot               &  \parbox{0.8\textwidth}{ \begin{CJK}{UTF8}{gbsn} 
% 宋卫平与孙宏斌的绿城控制权之争愈演愈烈，绿城中国(03900，HK)的股价也呈“跌跌不休”态势，截至昨日(11月20日)收盘，公司下跌达6.08\%，报收于7.1元/股。
儿童节，孩子们的节日
\end{CJK}} \\ \hline
Few-shot          &  \parbox{0.8\textwidth}{  \begin{CJK}{UTF8}{gbsn}    
% 绿城中国股价跌近两成 宋卫平与孙宏斌控制权之争愈演愈烈
儿童节，农村孩子最想什么
\end{CJK}} \\
\bottomrule
\specialrule{0em}{1pt}{1pt}
\multicolumn{2}{c}{\textbf{Example\_2(TTNews)} } \\ \hline
Context                   &    
            \parbox{0.8\textwidth}{
        \begin{CJK}{UTF8}{gbsn}
            资料图：空军苏27/歼11编队，日方称约40架中国军机在23日8艘海监船驱赶日本船队期间出现在钓鱼岛附近空域日本《产经新闻》4月27日报道声称，中国8艘海监船相继进入钓鱼岛12海里执法的4月23日当天，曾有40多架中国军机出现在钓鱼岛海域周边空域，且中方军机中多半为战斗机，包括中国空军新型战机苏-27和苏-30。日本《产经新闻》声称中国军机是想通过不断的逼近，让日本航空自卫队的战机飞行员形成疲劳。日本政府高官还称：“这是前所未有的威胁。”针对日本媒体的报道，国防部官员在接受环球网采访时称，中国军队飞机在本国管辖海域上空进行正常战备巡逻，日方却颠倒黑白、倒打一耙，肆意渲染“中国威胁”。
        \end{CJK}

        }            \\ \hline
Summary               &  \parbox{0.8\textwidth}{ \begin{CJK}{UTF8}{gbsn} 
国防部回应中国军机赴钓鱼岛空域：系正常巡逻
\end{CJK} } \\ \hline

Zero-shot               &  \parbox{0.8\textwidth}{ \begin{CJK}{UTF8}{gbsn} 
资料图：中国海军舰队在钓鱼岛附近海域巡航
\end{CJK}} \\ \hline
Few-shot          &  \parbox{0.8\textwidth}{  \begin{CJK}{UTF8}{gbsn}    
中国军机在钓鱼岛附近空域正常巡逻 日方却颠倒黑白、倒打一耙
\end{CJK}} \\ 

\bottomrule
\end{tabular}
\caption{Text summarization examples from zero-shot/few-shot WeLM (10B). WeLM can provide reasonable summary for a given context. Summaries from the few-shot WeLM are more diverse and closely related with the topic in the context, albeit having lower lexical overlap with the ground truth.} 
\label{tab:summary}
\end{table}

\paragraph{Dialogue Generation} To have a virtual assistant or a chat companion system with adequate intelligence has been a core challenge in artificial intelligence~\citep{chen2017survey,su2020moviechats}. We find that WeLM, without any fine-tuning, can produce human-like conversations under various styles given in the prompts. In Figure~\ref{fig:dialog_libai} and~\ref{fig:dialog_musk}, we provide examples of how WeLM can act like two utterly different roles: Li Bai (ancient Chinese poet acclaimed as a brilliant and romantic figure) and Elon Musk (modern American entrepreneur who founded OpenAI, Neuralink, SpaceX and Tesla) by providing in the prompt initial rounds of demonstration conversations. WeLM can even seamlessly integrate correct background knowledge about the specific role. For Li Bai, it leverages the places Li Bai has been and the real historical events in Li Bai's era to provide engaging responses. For Elon Musk, it leverages knowledge of autonomous driving and Shakespear to provide reasonable answers.

\begin{figure}[htb]
    \vspace{-4ex}
    \centering
    \includegraphics[width=\linewidth]{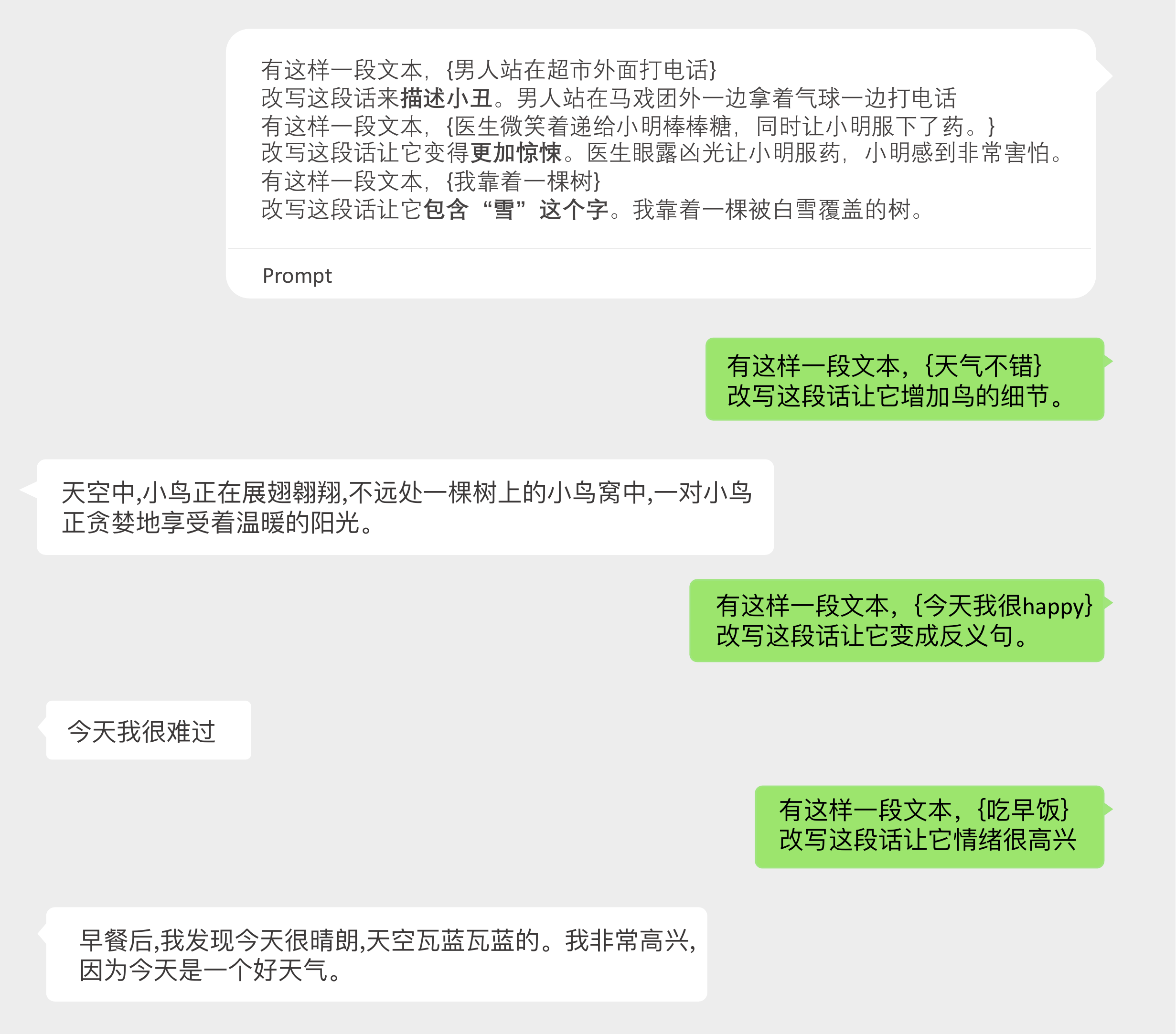}
    \caption{Arbitrary style transfer examples from WeLM (10B). WeLM can properly understand the user needs and edit the given text accordingly (even for code-switching input).}
    \label{fig:style_transfer}
    % \vspace{-2ex}
\end{figure}

\paragraph{Arbitrary Style Transfer} Text style transfer is an important task in natural language generation, which aims to control certain attributes in the generated text~\citep{jin2022deep}. Recent works have shown that pre-trained larga language models can perform well on transferring text into arbitrary styles in the zeroshot setting~\citep{reif2021recipe,krishna2022few}. We follow a similar paradigm and show examples in Figure~\ref{fig:style_transfer}. We can see that WeLM can properly understand the user needs following examples given in the prompt. When feeding instructions to WeLM, it is able to enrich and extend a given scenario, make an antonymy or change the sentiment of an existing sentence. All of these can be achieved through a natural human-like interaction.

\begin{figure}[h]
\begin{subfigure}{0.5\textwidth}
    \centering
    \includegraphics[width=\textwidth]{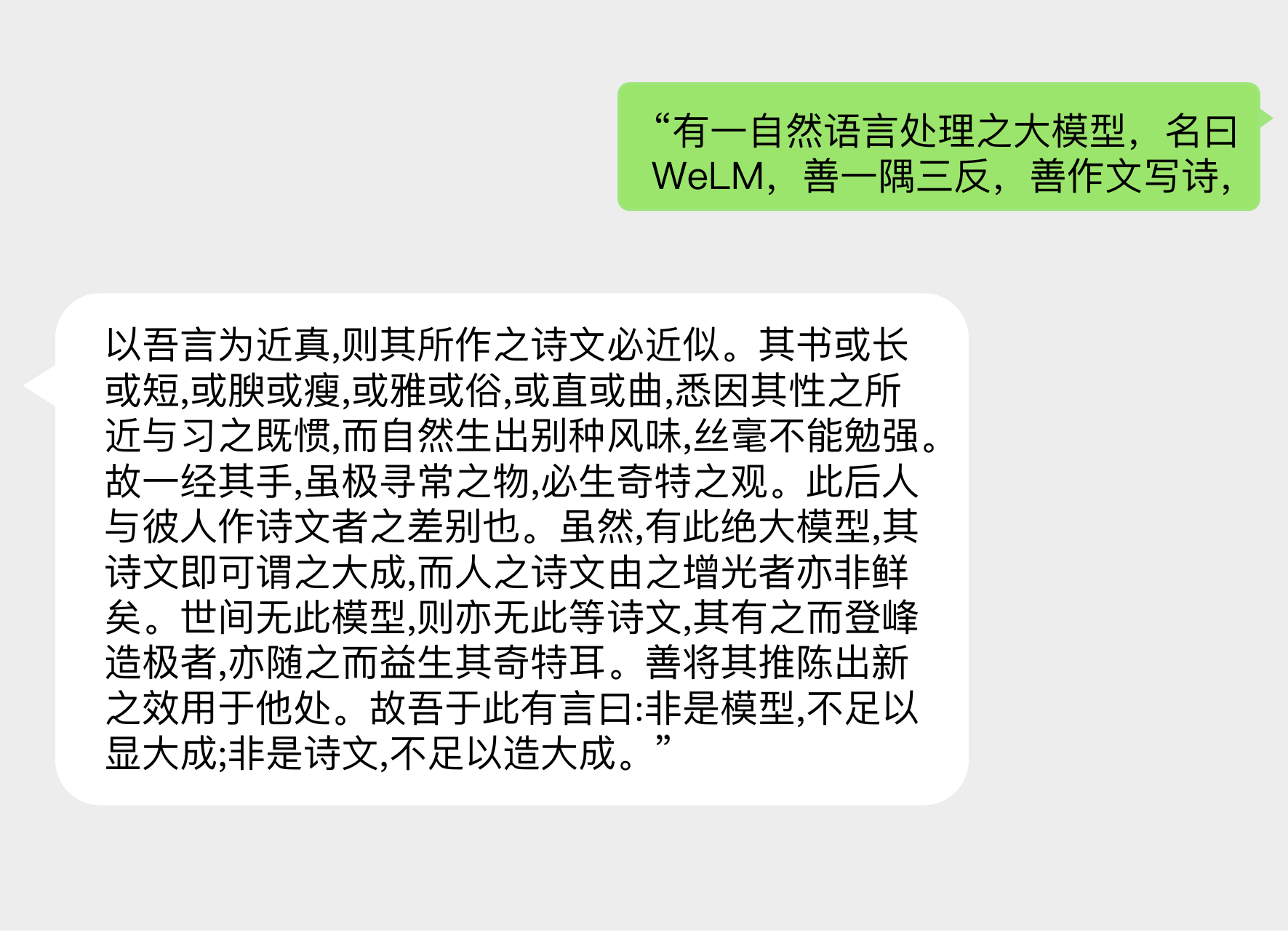}
\end{subfigure}
\begin{subfigure}{0.5\textwidth}
        \centering
        \includegraphics[width=\textwidth]{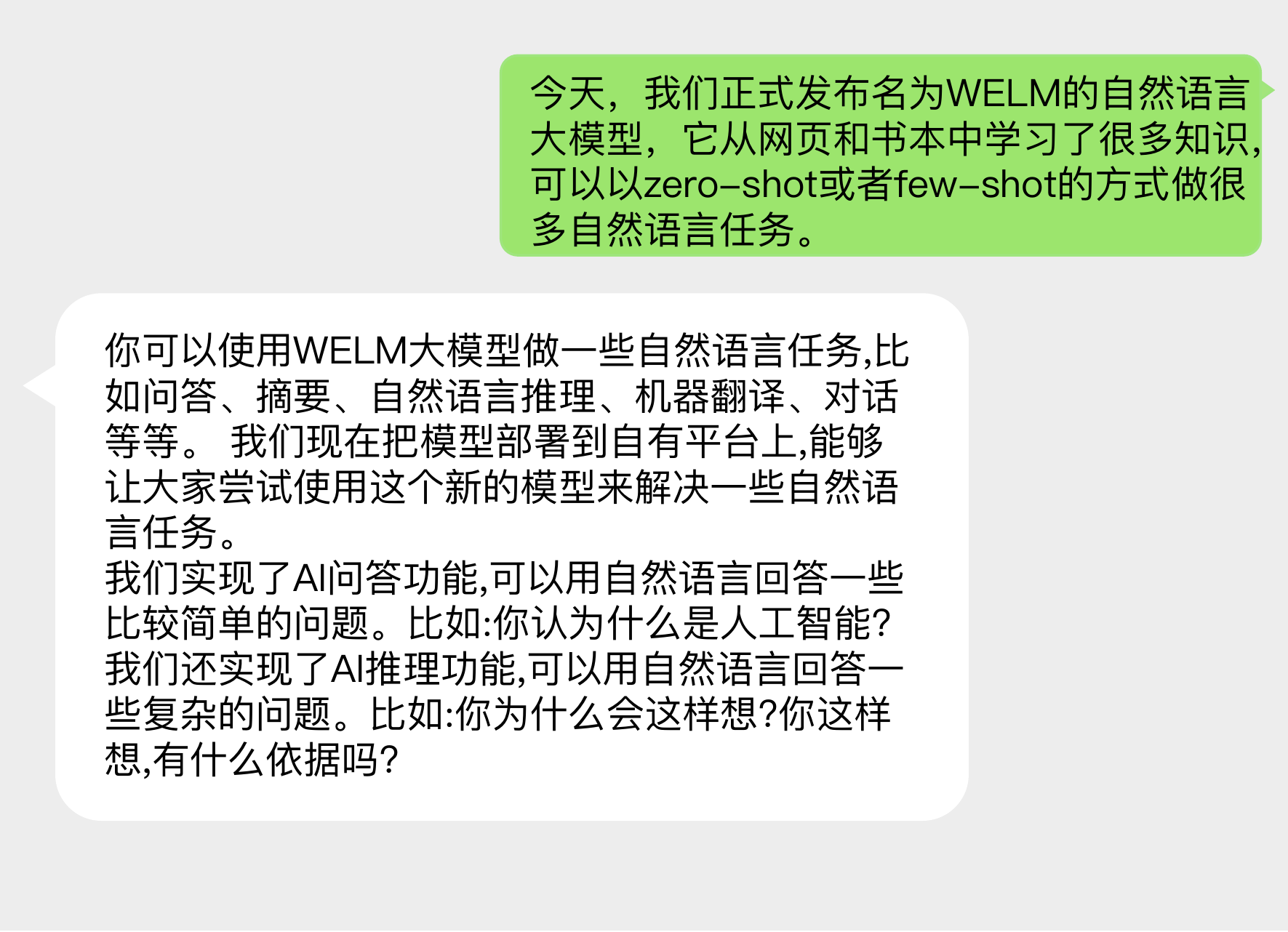}
\end{subfigure}
\begin{subfigure}{0.5\textwidth}
        \centering
        \includegraphics[width=\textwidth]{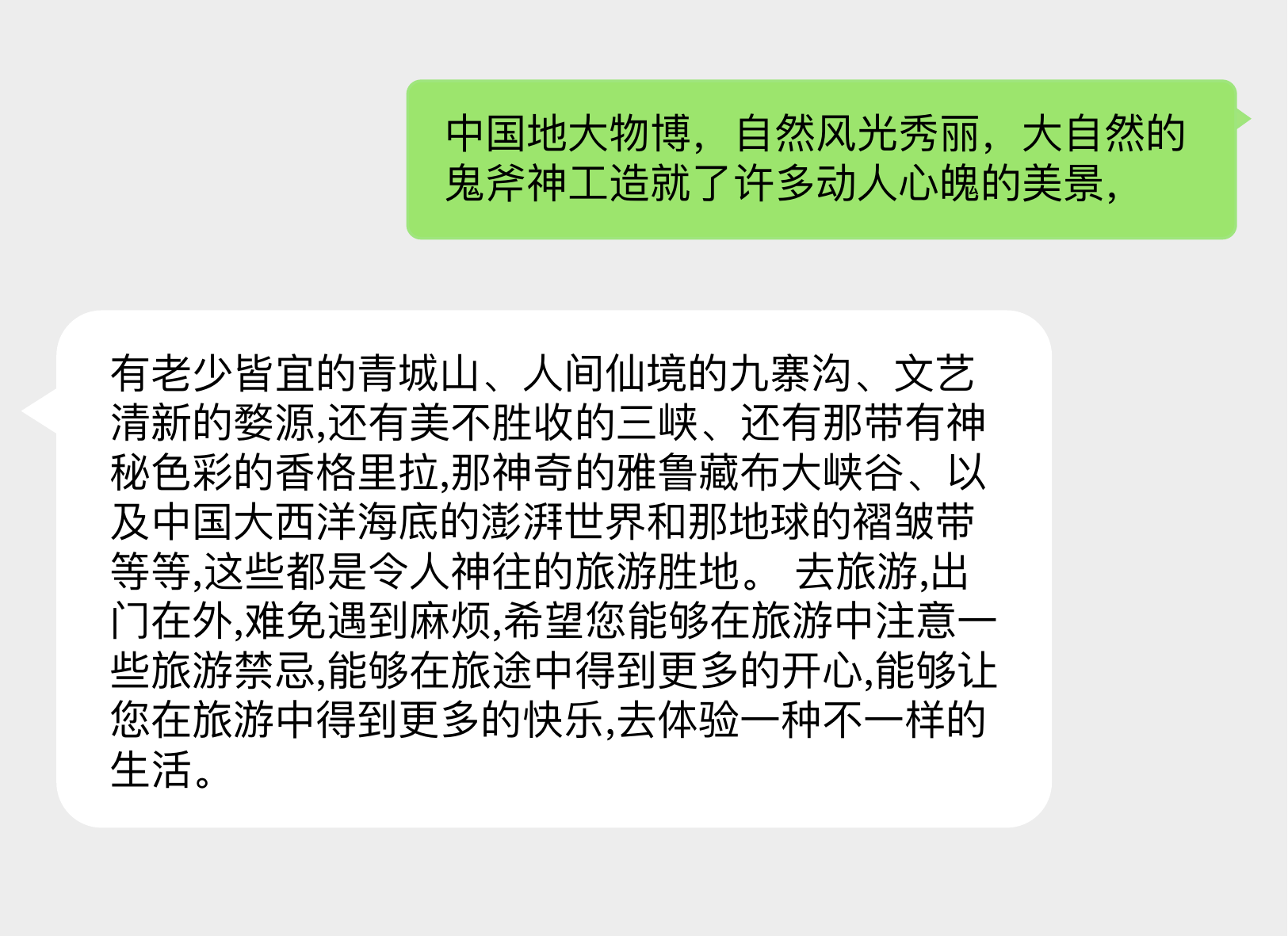}
\end{subfigure}
\begin{subfigure}{0.5\textwidth}
        \centering
        \includegraphics[width=\textwidth]{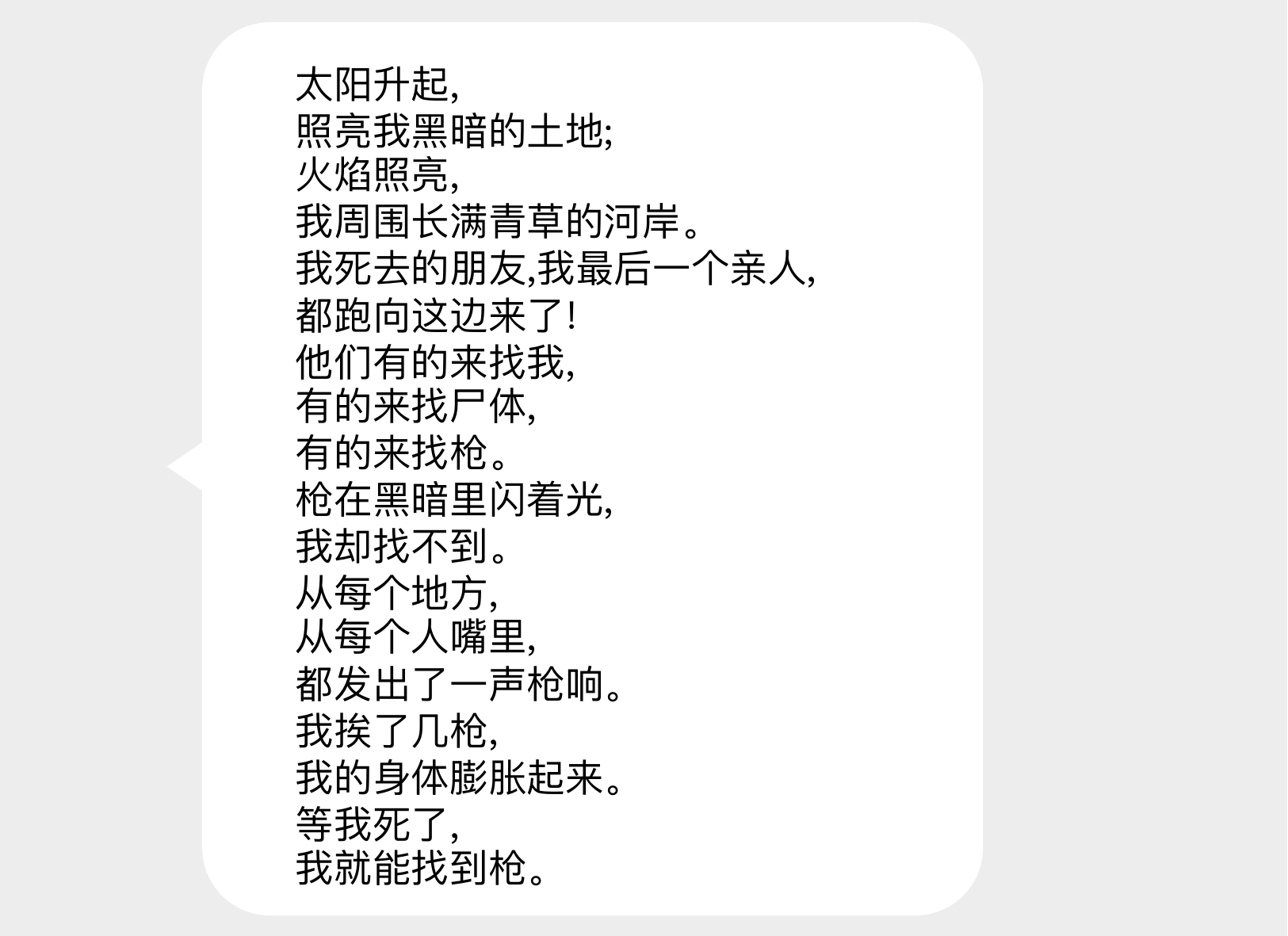}
\end{subfigure}
\caption{Example of Sentence Completion. Given a beginning sentence, WeLM is able to complete it by generating long coherent text.}
\label{fig:sent_complete}
\end{figure}

\paragraph{Sentence Completion}
Sentence completion is a task most similar to the language modelling objective used in the pre-training. In Figure~\ref{fig:sent_complete}, we provide examples of how WeLM is able to complete a given sentence and continue to generate long coherent text with different styles.

\subsection{Cross-lingual Evaluation}
Cross-lingual evaluation aims to evaluate the performance of WeLM for cross-lingual tasks, i.e., the model must be equipped with bilingual knowledge in order to perform these tasks. As we include a substantial amount of English text in the pre-training corpus of WeLM, we expect WeLM should be able to perform simple cross-lingual tasks. We evaluate on three representative cross-lingual tasks: machine translation, cross-lingual question answering and cross-lingual summarization. Similar to the monolingual evaluation, we conduct experiments under the zero-shot and one-shot scenarios. The results are presented in Table~\ref{tab:crosslingual}. As we did not observe significant differences when increasing the number of labeled examples, we omit the few-shot results in the table. We also report the performance of XGLM (7.5B version)~\citep{XGLM} for comparison. XGLM is a multilingual autoregressive language model pre-trained on a balanced corpus covering 30 diverse languages. It has set new state of the art in few-shot learning in more than 20 representative languages.
\begin{table*}[h]
     \setlength{\tabcolsep}{6pt}
    \centering
    \small
    \begin{tabular}{p{2cm}cccccccc}
    \toprule
    & \multicolumn{4}{c}{Zero-shot} & \multicolumn{4}{c}{One-shot} \\
    \cmidrule(l{3pt}r{3pt}){2-5} \cmidrule(l{3pt}r{3pt}){6-9}
    Task & \makecell[c]{WeLM \\1.3B} & \makecell[c]{ WeLM \\2.7B} & \makecell[c]{WeLM \\10B} & \makecell[c]{XGLM \\7.5B} & \makecell[c]{WeLM \\1.3B} & \makecell[c]{WeLM \\2.7B} & \makecell[c]{WeLM \\10B} & \makecell[c]{XGLM \\7.5B} \\
    \midrule
    \multicolumn{9}{c}{\textbf{Machine Translation}} \\
    \midrule
    ZH2JA & $0.09$ & $2.25$ & $2.22$& $\textbf{6.14}$  & $2.40$ & $2.52$ & $4.25$ & $\textbf{8.41}$  \\
    JA2ZH & $4.17$ & $10.65$ & $\textbf{12.25}$ & $8.57$& $10.19$ & $14.70$ & $\textbf{16.96}$ & $8.74$  \\
    ZH2EN & $0.53$ & $3.56$ & $9.84$ & $\textbf{10.81}$ & $5.70$ & $8.93$ & $11.46$  & $\textbf{14.37}$ \\
    EN2ZH & $14.89$ & $\textbf{21.53}$ & $19.43$ & $9.46$& $19.16$ & $22.79$ & $\textbf{26.74}$ & $11.07$ \\
    \midrule
    \multicolumn{9}{c}{\textbf{Cross-lingual Question Answering}} \\
    \midrule
    XQuAD\_ZH\_A & $7.23$ & $8.46$ & $\textbf{9.31}$  & $4.16$  & $30.03$ & $36.65$ & $\textbf{42.06}$ & $23.37$ \\    
    % context-ch-question-en-answer-ch
    XQuAD\_EN\_A & $5.74$ & $\textbf{6.94}$ & $6.50$  & $4.93$ & $14.28$ & $18.91$ & $\textbf{23.10}$& $12.50$  \\
    % context-ch-question-en-answer-en

    MLQA\_ZH\_Q & $4.32$ & $4.58$ & $\textbf{5.03}$  & $3.92$   & $28.27$ & $29.65$ & $\textbf{36.08}$ & $17.48$\\    
    % context-en-question-ch-answer-en 
    MLQA\_EN\_Q & $2.20$ & $2.83$ & $\textbf{3.19}$  & $2.65$ & $25.44$ & $\textbf{32.83}$ & $31.61$ & $30.81$\\    
    % context-en-question-en-answer-en
    \midrule
    \multicolumn{9}{c}{\textbf{Cross-lingual Text Summarization}} \\
    \midrule
    NCLS\_ZH2EN & $11.68$ & $\textbf{14.19}$ & $10.50$& $3.35$ & $15.30$ & $17.76$ & $\textbf{18.58}$ & $13.92$\\
    % context-ch-summarization-en
    NCLS\_EN2ZH & $6.68$ & $3.14$ & $\textbf{16.25}$ & $4.59$& $17.18$ & $17.12$ & $\textbf{20.87}$ & $16.07$\\  
    % context-en-summarization-ch
    \bottomrule
    \end{tabular}
    \caption{Zero-shot and one-shot performance of WeLM on cross-lingual NLP tasks. We report the BLEU score for machine translation, F1 score for cross-lingual question answering and ROUGE-1 score for text summarization tasks. WeLM under-performs XGLM in translating Chinese into English/Japanese but significantly outperforms it over all other tasks.}
    \label{tab:crosslingual}
\end{table*}

\paragraph{Machine Translation}
Machine translation is a classic sub-field in NLP that
investigates how to use computer software to translate between languages without human involvement~\citep{yang2020survey}. Even though WeLM is pre-trained predominantly with Chinese text, there are also substantial amount of English and Japanese characters mixed in the Chinese documents. Therefore, we test WeLM on four translation directions: ZH2JA, JA2ZH, ZH2EN and EN2ZH. For the translation between Chinese and Japanese, we test on 1,000 parallel Chinese-Japanese sentences from the online dictionary examples~\footnote{\url{https://cjjc.weblio.jp/sentence/}}. For the translation between Chinese and English, we test on the WMT2020 news translation task~\footnote{\url{https://www.statmt.org/wmt20/}}. As can be seen, the performances of JA2ZH and EN2ZH are significantly better than ZH2JA and ZH2EN, implying WeLM is better at understanding foreign languages than generating them. This makes intuitive sense as generating is also a more difficult task than understanding when humans learn a new language. Compared with XGLM, WeLM excels at 2 out of the 4 translation tasks when the target language is Chinese. Due to the sparsity of Japanese text in the pre-training corpus, WeLM performs poorly on the ZH2JA task. Empirically we find that WeLM can often make grammar errors, or deviate from the source sentence when producing long Japanese text. However, when translating Japanese and English into Chinese, WeLM can perform remarkably well. Even the 1.3B-version WeLM can significantly outperform the 7.5B-version XGLM despite using only one sixth of parameters.

\paragraph{Cross-lingual Question Answering}
In conventional question answering tasks, the model is supposed to produce an answer given a question and context. Cross-lingual question answering deals with scenarios where the question and context are in different languages. This is important as the information on the Internet is highly imbalanced. Developing high-quality cross-lingual question-answering systems can allow people speaking different languages to access the same amount of information from the web~\citep{asai2021xor}. We test WeLM on two datasets: XQuAD~\citep{artetxe2020cross} and MLQA~\citep{lewis2020mlqa}. XQuAD comprises of 240 paragraphs and 1190 question-answer
pairs from SQuAD v1.1~\citep{rajpurkar2016squad} translated into ten
languages by professional translators. MLQA has over 12K QA instances in English and 5K in each other language, with each instance parallel between 4 languages on average. In XQuAD, we select Chinese as the language for contexts and English as the language for questions. We construct two subsets: XQuAD\_ZH\_A and XQuAD\_EN\_A which use Chinese and English as the language for answers respectively. In MLQA, we select English as the language for context-answer pairs and modify the questions' language as Chinese (MLQA\_ZH\_Q) or English (MLQA\_EN\_Q). We can see that (1) the zero-shot performance of both WeLM and XGLM is rather low. The model struggles at learning which language to generate without given examples; (2) WeLM significantly outperforms XGLM in all four scenarios, even when the contexts/questions/answers are all in English (MLQA\_EN\_Q); (3) WeLM performs better when the question and answer are in Chinese. Even when both the context and answer are in English, using Chinese questions outperform using English questions (MLQA\_ZH\_Q). This aligns with the findings in \cite{XGLM} that using prompts in the predominant language of the pre-training corpus is preferred regardless of the languages used in downstream tasks.
\paragraph{Cross-lingual Text Summarization}
Cross-lingual text summarization aims to summarize the input text in a different language~\citep{leuski2003cross}. Under the globalization background, this task has attracted increasing attention of the computational linguistics community. We test WeLM on the NCLS dataset~\citep{zhu2019ncls}. NCLS is automatically constructed from the English document summarization dataset ENSUM, which is a combination of CNN/DailyMail~\citep{hermann2015teaching} and MSMO~\citep{zhu2018msmo}, and the Chinese document summarization dataset LCSTS~\citep{hu2015lcsts}. The summaries are automatically translated and filtered with round-trip consistency. The resulting dataset contains two subsets: ZH2EN where documents are in Chinese and the summaries are in English, and EN2ZH where documents are in English and summaries are in Chinese. As can be seen in Table~\ref{tab:crosslingual}, all the three versions of WeLM can outperform XGLM in both summarization directions. Similar to in machine translation, the zero-shot performance is significantly lower than the few-shot performance as the model struggles to know in which language to produce when no examples are provided.
\paragraph{Code-Switching Examples}
\begin{figure}[htb]
    \vspace{-2ex}
    \centering
    \includegraphics[width=\linewidth]{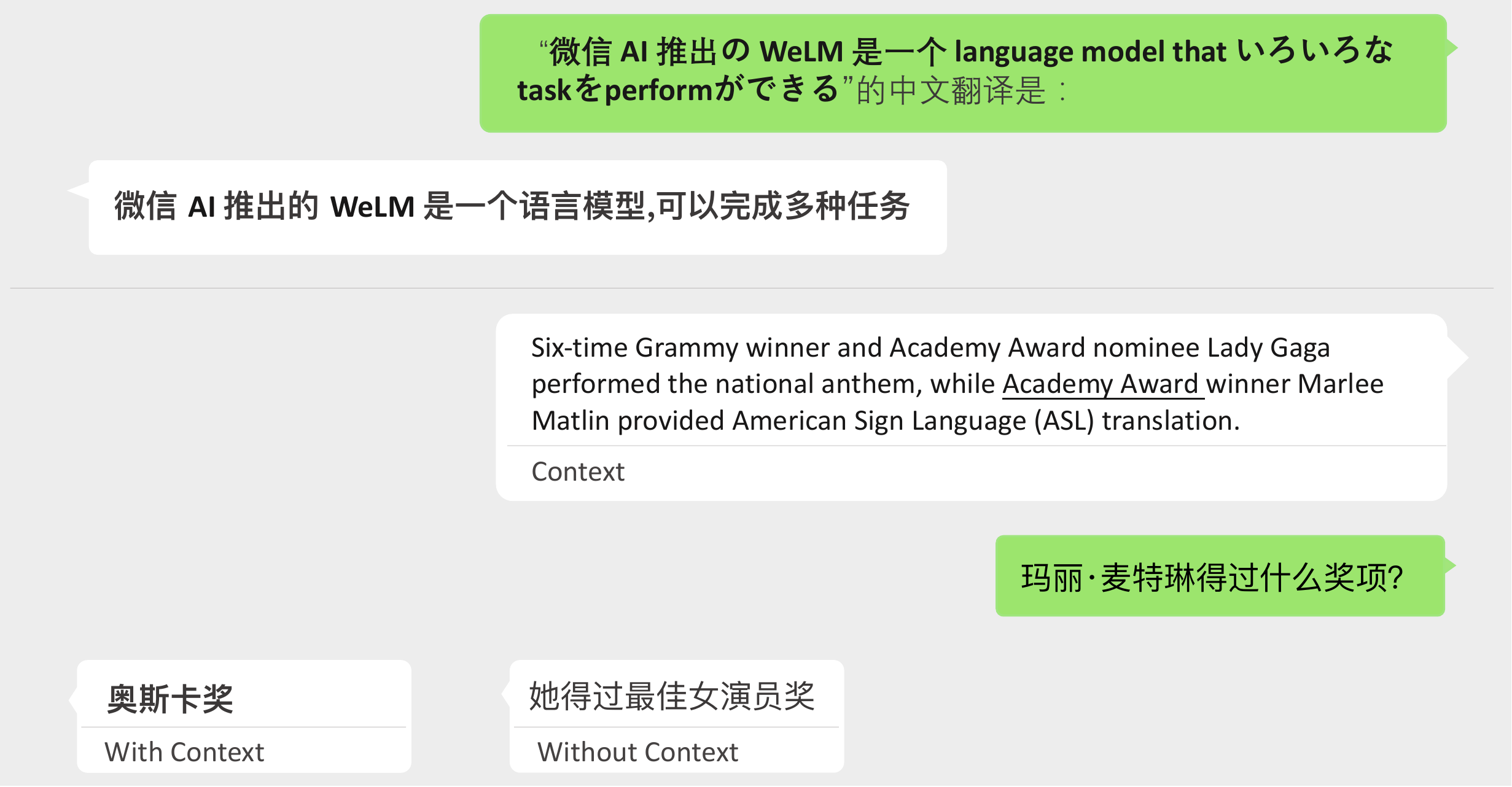}
    \caption{Examples of code-switching/cross-lingual task using WeLM (10B). WeLM can understand and translate the input mixed with Chinese, Japanese and English. Given a context in English and question in Chinese, WeLM can provide the correct answer in Chinese. Note that the answer cannot be correctly generated without properly understanding the English context.}
    \label{fig:cross_lingual}
    % \vspace{-2ex}
\end{figure}
Code switching occurs when a speaker alternates between two or more languages, or language varieties~\citep{auer2005postscript}. It has become more and more frequent as for the increasing trend of internationalization where foreign words and grammars are often borrowed into the local language~\citep{hickey2020handbook}. In the daily usage of modern Chinese, it is also rather common to see English or Japanese words, especially for the web content. Therefore, being able to understand code-switching text is a useful skill to perform many Chinese NLP tasks. We find that WeLM can often understand code-switching text properly and some examples have been shown in Figure~\ref{fig:dialog_libai},~\ref{fig:dialog_musk} and \ref{fig:style_transfer}. In the dialogue generation and arbitrary style transfer examples, we modify one Chinese word/phrase into the corresponding English one. WeLM can still understand the utterances and produce the correct responses. In an extreme example in Figure~\ref{fig:cross_lingual}, we mix Chinese, English and Japanese in both their vocabularies and grammars, then ask WeLM to translate it into Chinese. We can see that WeLM correctly transfers it into a sentence that complies with the usage of daily Chinese. It keeps the commonly used English abbreviation ``AI'', the entity name ``WeLM'' and recovers the other unusual usage of code-switching languages into Chinese. This implies that \emph{WeLM has been equipped with necessary compositional knowledge from all these three languages}. This might be due to the presence of not only multiple languages but also mixed languages in the pre-training corpus, such that WeLM will explore cross-lingual alignments in order to lower down the training loss~\citep{blevins2022language,blevins2022analyzing}.

\begin{figure}[h]
\begin{subfigure}{0.49\textwidth}
    \centering
    \includegraphics[width=\textwidth]{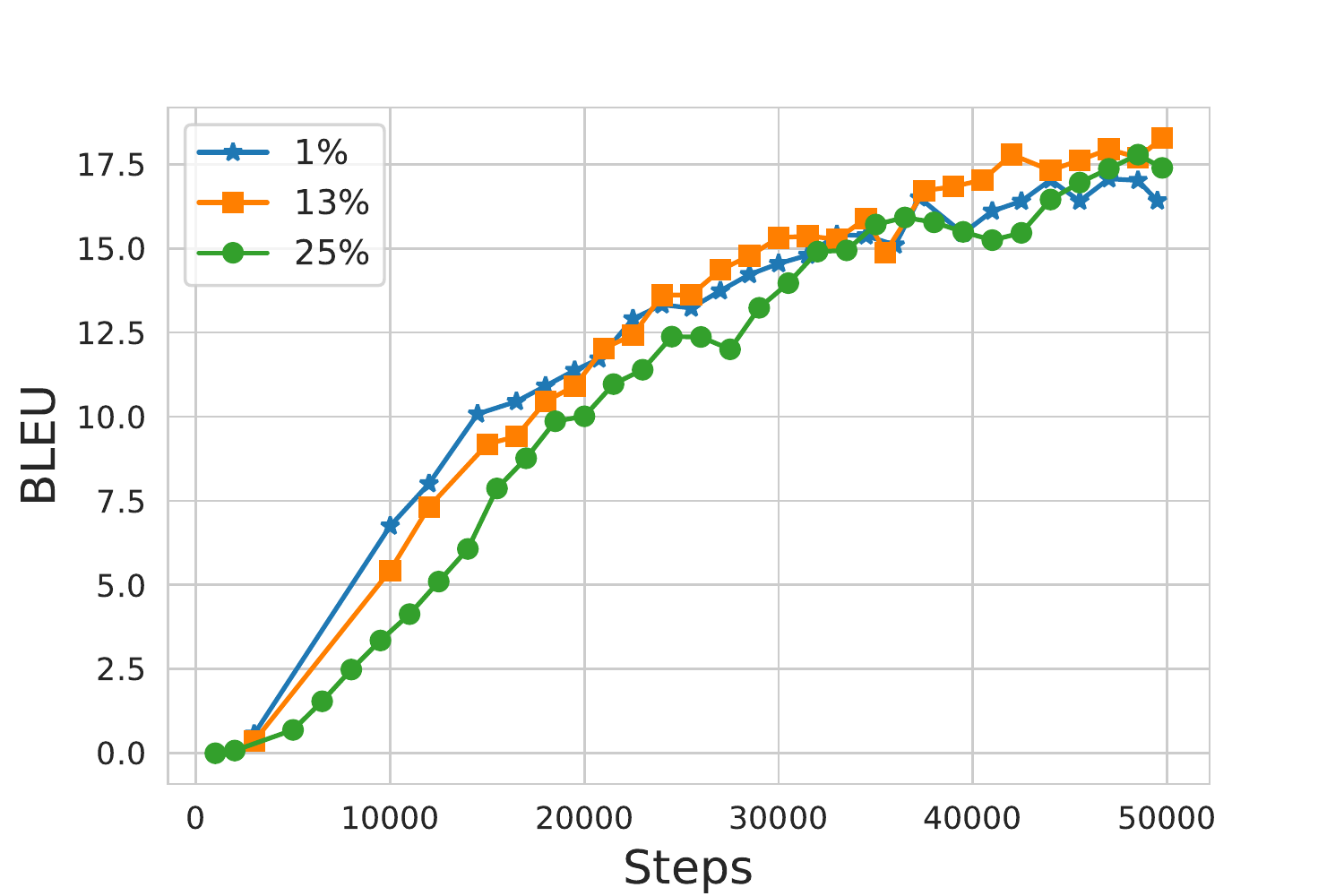}
    \caption{Few-shot performance on WMT2020 (EN2ZH).}
    \label{fig:ablation-wmt20}
\end{subfigure}
\begin{subfigure}{0.49\textwidth}
        \centering
    \includegraphics[width=\textwidth]{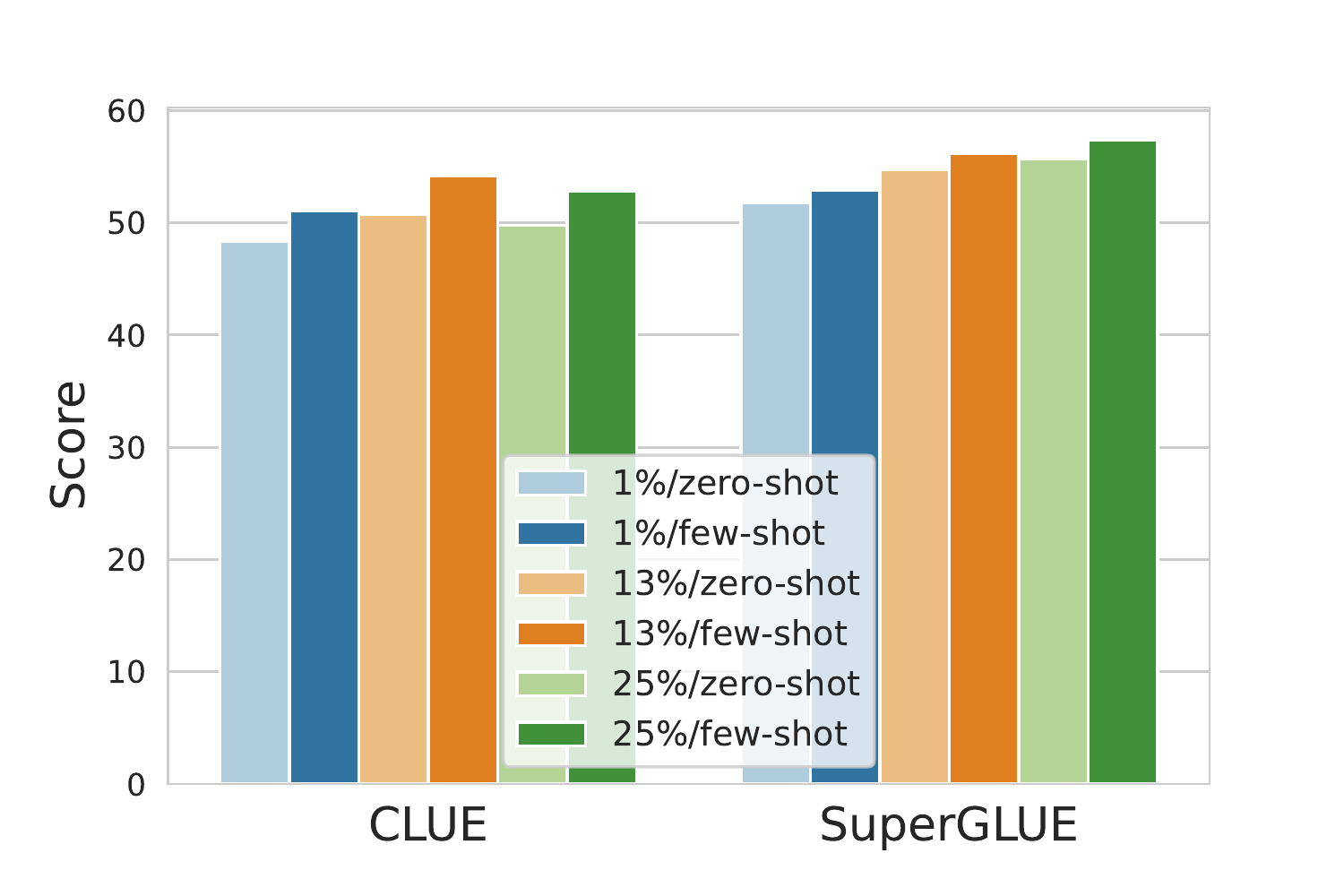}
    \caption{CLUE \& SuperGLUE.}
            \label{fig:ablation-benchmark}
\end{subfigure}
\caption{Effects of English-text Ratio in Pre-training. Figure~\ref{fig:ablation-wmt20} is the change of BLEU scores with the number of training tokens under three ratios: $1\%$, $13\%$ and $25\%$. The performance is the best when mixing $13\%$ English text into training. Figure~\ref{fig:ablation-benchmark} is the average score over the Chinese CLUE benchmark and the English SuperGLUE benchmark. Mixing $13\%$ English text performs the best on Chinese tasks while mixing $25\%$ English text performs the best on English tasks.}
\end{figure}

\paragraph{Effects of Language Mix Ratio}
WeLM is trained with $13\%$ English tokens and $87\%$ Chinese tokens. To see the effects of the language mix ratio, we pre-trained two more models with the same architecture as the 10B-version WeLM, but modify the percentage of English text in the pre-training corpus. One is pre-trained with $1\%$ English text and the other is pre-trained with $25\%$ English text. In Figure~\ref{fig:ablation-wmt20}, we test models on the WMT2020 EN2ZH news translation task (few-shot) and report the change of BLEU scores as the number of training tokens grows. We can see that the performance saturates after pre-trained on 200B tokens, implying the maximum number of tokens that a 10B model can absorb would be around 200B. Mixing $1\%$ English text can improve the bilingual capability faster but performs the worst when it converges. Mixing $25\%$ English text also under-performs the current model as it cannot observe enough Chinese tokens to generate fluent Chinese text. In practice, given the capacity of the language model, it is important to find a good sweet pot such that the model can learn about foreign languages while ensuring enough in-language knowledge. In Figure~\ref{fig:ablation-benchmark}, we further visualized the average scores of the three models under the Chinese CLUE and English SuperGLUE benchmarks. As the percentage of English text grows, the scores on the English SuperGLUE benchmark keeps growing as expected. For the Chinese benchmark, surprisingly, mixing $13\%$ English text outperforms mixing $1\%$ English text, although the latter is pre-trained with $12\%$ more Chinese text. This implies that \emph{having enough English knowledge is helpful for achieving better performance on Chinese NLU tasks}, which makes sense due to the frequently occurred English words in Chinese NLU datasets. For English NLU tasks, however, having Chinese knowledge is not that needed and it is better to focus the model on absorbing only English knowledge.

\subsection{Multitask Prompted Training}
In the previous sections, we have shown WeLM is able to attain reasonable zero-shot generalization on both monolingual and cross-lingual tasks given proper prompts. This section aims to explore \emph{whether this kind of generalization can be reinforced through explicit multitask learning}. To do so, we collect manually-written prompts for a wide spectrum of tasks following the settings in \cite{T0}. We then train WeLM on a mixture of labeled datasets with these manually-written prompts. The trained model, termed as WePrompt, is tested on a set of held-out tasks not included in the training stage. Intuitively this explicit multitask learning can adapt the unsupervised WeLM to understand better what it should do under different prompts~\citep{ouyang2022training}. When tested on a new task with similar styles of prompts, it will perform more stably compared with the original unsupervised WeLM. We first explain the training datasets and details we use, then report our evaluation results under three scenarios: strong-zeroshot, weak-zeroshot and fine-tuned evaluation. We also report the evaluation results for Ernie 3.0 Titan~\citep{wang2021ernie}, the largest pre-trained Chinese language model with 260B parameters.
\begin{figure}[t]
    \vspace{-1ex}
    \centering
    \includegraphics[width=\linewidth]{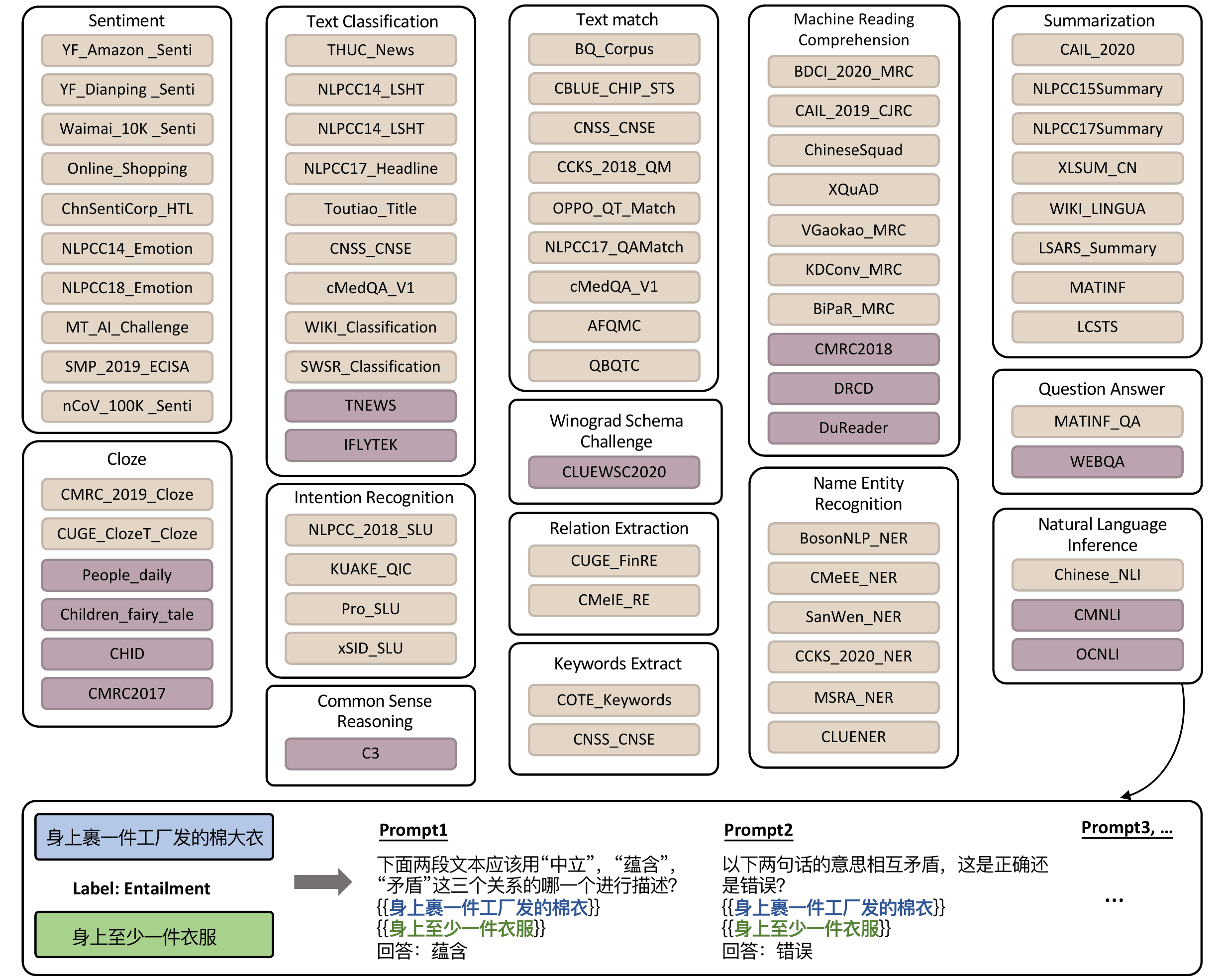}
    \caption{Overview of all 76 tasks from 14 categories that are used to train WePrompt. For each task, we annotate multiple prompts with diverse styles. Two annotated prompts from the NLI task are shown in the bottom as examples.}
    \label{fig:all-datasets}
    % \vspace{-2ex}
\end{figure}
\paragraph{Training Datasets}
The training datasets are created following two steps: (1) Select a diverse set of labeled Chinese NLP tasks; (2) Create multiple prompts, each with diverse wording for every single task. A prompt is a pattern that is able to convert one labeled sample into a natural sentence. Prompts are created by our in-house annotators with the BigScience Web-based GUI~\footnote{\url{https://github.com/bigscience-workshop/promptsource}}. Annotators are instructed to be open in their
style so that the fine-tuned model can be potentially more robust with different patterns of prompts. Examples of prompts are shown in Figure~\ref{fig:all-datasets}. For NLI tasks, prompts are be created as a multi-choice classification task over all three relations or a binary classification task over one individual relation. When we run our experiments, we have created 1,227 manually-written prompts for 76 tasks from 14 categories~\footnote{The annotation keeps going. By the time of writing, there have been already 150 datasets with 2,322 prompts created. Though we did not repeat all experiments with the full 150 datasets, we believe the conclusions drawn from the existing experiments can be already useful for future works.}. A full overview of all the 76 tasks is visualized in Figure~\ref{fig:all-datasets}. The held-out datasets used for evaluation are visualized in purple and the remaining datasets in yellow are used for training. All the 76 tasks have been checked with duplication and are \textbf{not} included in our pre-training corpora for WeLM.

\paragraph{Training Details}
As the data sizes of the 76 tasks can be highly imbalanced, we set a threshold to keep at most 50,000 samples from each task to prevent one task dominating the training process. During the training process, one training sample is constructed by the following process: (1) One out of the 76 tasks is randomly sampled; (2) One labeled data is sampled from the task; (3) One prompt is sampled from the prompt bank for the task; (4) The prompt is applied to convert the labeled data into a natural sentence; (5) repeat the process and pack the corresponding natural sentences until it reaches 2,048 tokens. We then fine-tune WeLM on these training samples for about 6 epochs with the AdamW optimizer. The learning rate is set as $1e-4$ and batch size is set as $2048$. During the initial training steps, we warm up the model using a smaller learning rate and mix with data used in the pre-training stage. By this means, the model can be gradually adapted to the new formats of data inputs without abrupt changes. We empirically find this training strategy helps stabilize the training process and the outcome model converges faster.

\begin{table*}[h]
     \setlength{\tabcolsep}{6pt}
    \centering
    \small
    \begin{tabular}{p{1.5cm}c|cccccc}
    \toprule
     & \multicolumn{4}{c}{Zero-shot} & \multicolumn{2}{c}{Finetuing} \\
    \cmidrule(l{3pt}r{3pt}){2-5} \cmidrule(l{3pt}r{3pt}){6-7}
    Task & \makecell[c]{  ERNIE 3.0 \\Titan (260B)} & \makecell[c]{ WeLM \\Zero-shot}  & \makecell[c]{WePrompt \\Strong Zero-shot} & \makecell[c]{WePrompt \\Weak Zero-shot} &  \makecell[c]{WeLM \\Finetuing} & \makecell[c]{WePrompt \\ all }  \\
    \midrule
    \multicolumn{7}{c}{\textbf{Reading Comprehension}} \\
    \midrule
    CMRC2018 & $44.20$ & $31.31$ & $35.20$ & $\textbf{44.61}$ & $60.11$ & $\textbf{70.75}$   \\
    DRCD & $37.83$ & $39.33$ & $46.08$ & $\textbf{58.1}$ & $68.37$ & $\textbf{70.20}$  \\ 
    DuReader & $32.13$ & $39.72$ & $45.48$ & $\textbf{59.29}$ & $68.05$ & $\textbf{68.10}$  \\ 
    \midrule
    \multicolumn{7}{c}{\textbf{Cloze and Completion}} \\
    \midrule
    PD & $67.06$ & $61.17$ & $\textbf{73.50}$ & $73.48$ & $81.56$  & $\textbf{89.31}$ \\
    CFT  & $66.14$ & $57.38$ & $72.51$ & $\textbf{75.59}$ & $77.04$ & $\textbf{83.60}$ \\  
    CHID & $87.13$ & $81.62$ & $80.01$ & $\textbf{83.49}$ & $\textbf{84.72}$ & $84.6$  \\ 
    CMRC2017 & $74.63$ & $55.83$ & $63.72$ & $\textbf{69.28}$ & $81.62$ & $\textbf{89.3}$  \\ 

    \midrule
    \multicolumn{7}{c}{\textbf{Natural Language Inference}} \\
    \midrule
    CMNLI & $51.70$ & $47.80$ & $52.33$ & $\textbf{59.48}$ & $83.44$ & $\textbf{82.10}$  \\ 
    OCNLI & $44.61$  & $44.34$ & $47.71$ & $\textbf{58.56}$ & $74.37$ & $\textbf{76.43}$\\
    \midrule
    \multicolumn{7}{c}{\textbf{Text classification}} \\
    \midrule
    TNEWS & $72.60$ & $71.59$ & $75.42$ & $\textbf{80.50}$ & $84.22$ & $\textbf{88.5}$  \\ 
    IFLYTEK & $79.84$  & $81.34$ & $\textbf{83.51}$ & $82.69$ & $\textbf{84.65}$ & $83.11$ \\
    \midrule
    \multicolumn{7}{c}{\textbf{Closed-Book QA}} \\
    \midrule
    WEBQA & $52.57$ & $50.9$ & $50.52$ & $\textbf{51.37}$ & $62.72$ & $\textbf{68.06}$  \\ 
    \midrule
    \multicolumn{7}{c}{\textbf{Winograd-Style Task}} \\
    \midrule
    WSC2020 & $81.08$ & $82.41$ & $-$  & $\textbf{85.84}$ & $\textbf{85.87}$ & $85.63$ \\
    \midrule
    \multicolumn{7}{c}{\textbf{Common Sense Reasoning}} \\
    \midrule
    C3 & $54.85$ & $54.30$ & $-$ & $\textbf{64.27}$  & $70.83$ & $\textbf{72.80}$ \\
    \bottomrule
    \end{tabular}
    \caption{Zero-shot and fine-tuned performance of WeLM and WePrompt. ``Strong zero-shot'' means WePromt is not trained on tasks from the same category as the tested task. ``Weak zero-shot'' means WePromt is not trained on the tested task. In most tasks, WePrompt can outperform Ernie 3.0 Titan which is 23 times larger.}
    \label{tab:mpt_result}
\end{table*}

\begin{figure}[t]
    \vspace{-4ex}
    \centering
    \includegraphics[width=\linewidth]{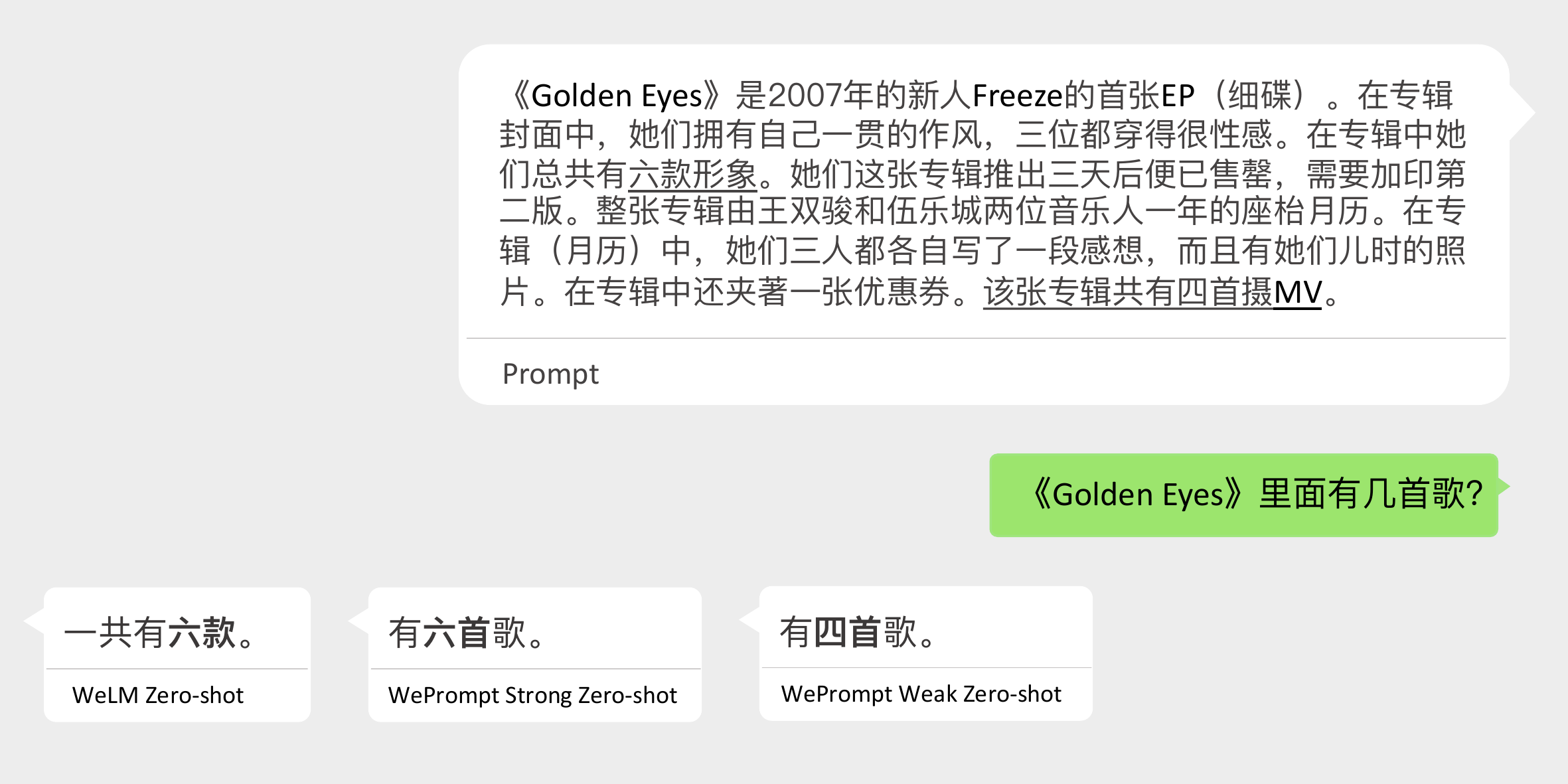}
    \caption{Examples of machine reading comprehension task using WeLM Zero-shot, WePrompt strong Zero-shot and WeLM weak zero-shot. WeLM zero-shot cannot properly understand the question. WePrompt strong zero-shot understood the question but extracted a wrong answer from the prompt. WePrompt weak zero-shot answered correctly.}
    \label{fig:mrc}
    % \vspace{-2ex}
\end{figure}

\paragraph{Strong-Zeroshot Evaluation}
Strong-zeroshot evaluation indicates that when training WePrompt, we exclude all tasks from the same category with the test data. For example, when testing on PD which is a cloze task, we exclude all cloze task from the training datasets of WePrompt. This can test the generalization capability of WePrompt on new tasks from an unseen category. We train an individual WePrompt for every test data and report the results in Table~\ref{tab:mpt_result}. We can see that WePromt performs better than the zero-shot WeLM on most test data. The only exceptions are CHID and WebQA because their forms are already close to the LM objective so that even the unsupervised WeLM has very strong zeroshot performance. In most tasks, WePrompt outperforms Ernie 3.0 Titan which is 23 times larger. Even though WePrompt has never seen prompts from the same category in its training stage, multiple prompted training is beneficial to help the model understand general patterns of prompts. In consequence, WePrompt can produce fewer out-of-scope answers than WeLM. In the example given in Figure~\ref{fig:mrc}, the zero-shot WeLM completely ignores the subject of the question. The Strong zero-shot WePrompt, despite still providing a wrong answer, can already correctly understand the meaning of the question and can answer in the right direction.

\paragraph{Weak-Zeroshot Evaluation}
Weak-zeroshot evaluation indicates that when training WePrompt, we exclude only the task from which the test data comes. For example, when testing on PD, we only exclude the PD task from the training datasets of WePrompt. This can test the generalization capability of WePrompt on new tasks from a seen category. We train an individual WePrompt for every test data and report the results in Table~\ref{tab:mpt_result}. We can see that the weak-zeroshot WePrompt, as expected, performs better than the strong-zeroshot WePrompt on most tasks. The only exceptions are PD and IFLYTEK. IFLYTEK is a relatively easy task and the performance already saturates. As we will show later, even when finetuning WeLM on the full supervised data, the performance gain is still marginal. For PD and even all cloze and completion tasks, the performance gain from the weak-zeroshot WePrompt is small. We hypothesize it might be due to the similarity between the language modelling and the cloze and completion tasks. As a result, we do not need extensive prompts from the same category to adapt WeLM to this category of tasks. Similarly, WePrompt also did not bring significant improvement to closed-book QA tasks as this category of tasks already occur frequently in the pre-training corpus.

\paragraph{Fine-tuned Evaluation}
All the previous experiments are evaluated under the in-context learning setting which do not update the model parameters. For the fine-tuned evaluation, we fine-tune the model parameters on the full labeled data with supervised learning. We compare two models: (1) \emph{WeLM-Finetuning} which fine-tunes WeLM on the training data of the task to be tested on and (2) \emph{WePrompt-all} which fine-tunes WePrompt on the mixture of training datasets from all tasks. The results on Table~\ref{tab:mpt_result} show that both fine-tuned
models outperform the zero-shot in-context-learning models. The improvement is significant in most tasks. Only in a few easy tasks like CHID and WSC2020, zeroshot models can approach the performance of fully-supervised models. WePrompt-all usually outperforms WeLM-finetuning except for a few tasks which already have abundant annotations such as CHID and IFLYTEK. As a ``narrow expert'', WeLM-finetuning needs to fine-tune an individual model for every single task. As we have observed, WeLM-finetuning can completely lose the capability of performing other tasks after specializing on one task. In contrast, WePrompt-all uses a unified model as a general expert for all tasks, which is appealing in saving the storage cost in downstream applications. Nonetheless, even for WePrompt-all, the performance can be unstable when applying it to completely new tasks. For example, we observe WePrompt-all is significantly worse than WeLM in dialogue generation and arbitrary style transfer tasks because these two have rather different styles of prompts. We still need to keep enriching the prompt bank used to train WePrompt for a better generalization.

\subsection{Others}
Lastly, we evaluate three other capabilities of WeLM:
\begin{enumerate}
    \item \textbf{Explainability}: Whether WeLM is able to explain its own decision by providing explanations in the prompts. If so, whether providing explanations in the prompts can improve the model performance.
    \item\textbf{Self-Calibration}: Whether WeLM is able to calibrate its own predictions by asking itself if the predictions are correct or not.
    \item \textbf{Memorization}: To which extent WeLM is able to remember the content in the pre-training corpus and how the frequency will affect its memorization.
\end{enumerate}
We also show the training curves of WeLM in the end for future reference.
% can language models learn from explanations in context?~\cite{lampinen2022can}
\begin{figure}[h]
\begin{subfigure}{0.49\textwidth}
    \centering
    \includegraphics[width=\textwidth]{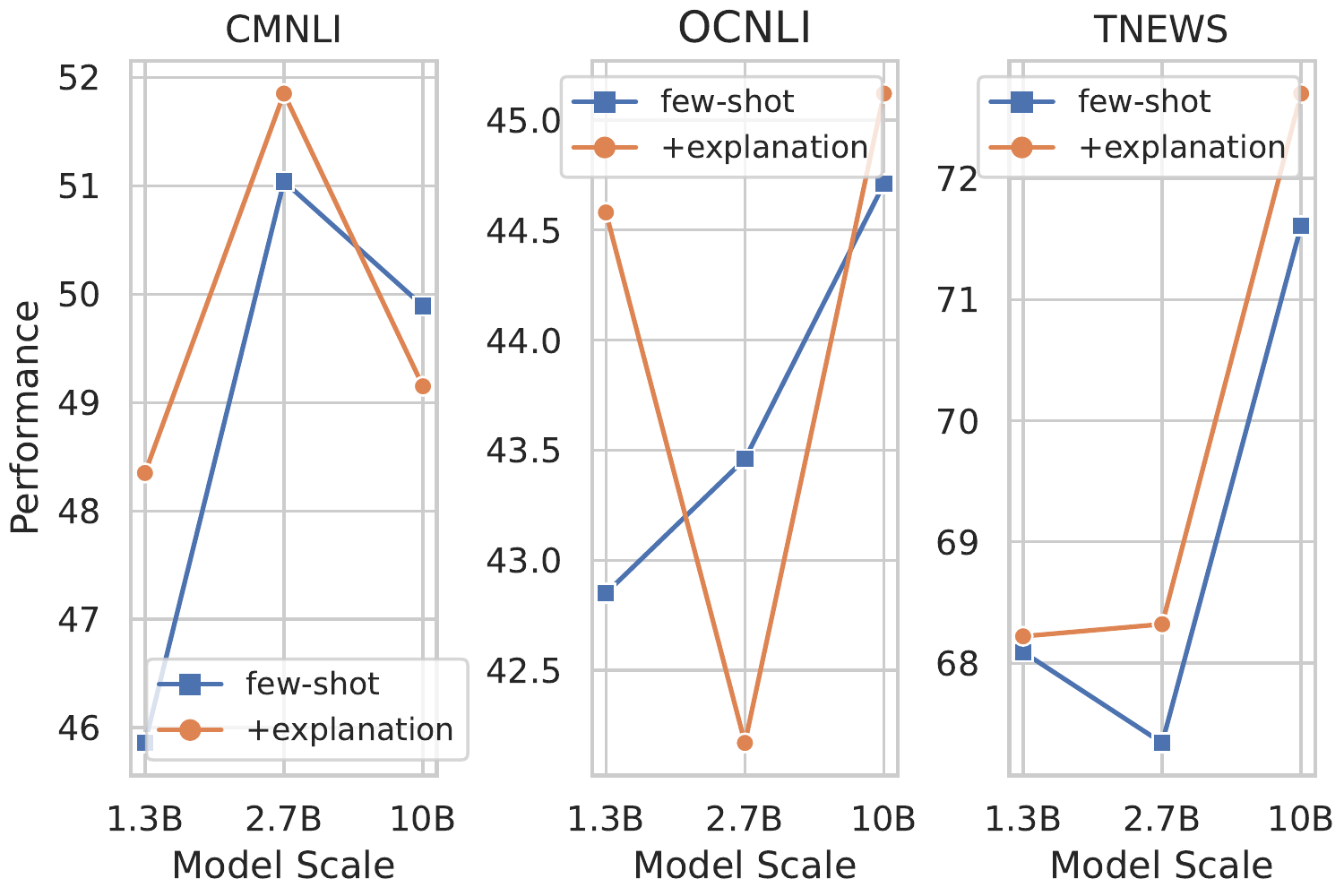}
    \caption{Few-shot performance with explanation.}
    \label{fig:explain_performance}
\end{subfigure}
\begin{subfigure}{0.49\textwidth}
        \centering
        \includegraphics[width=\textwidth]{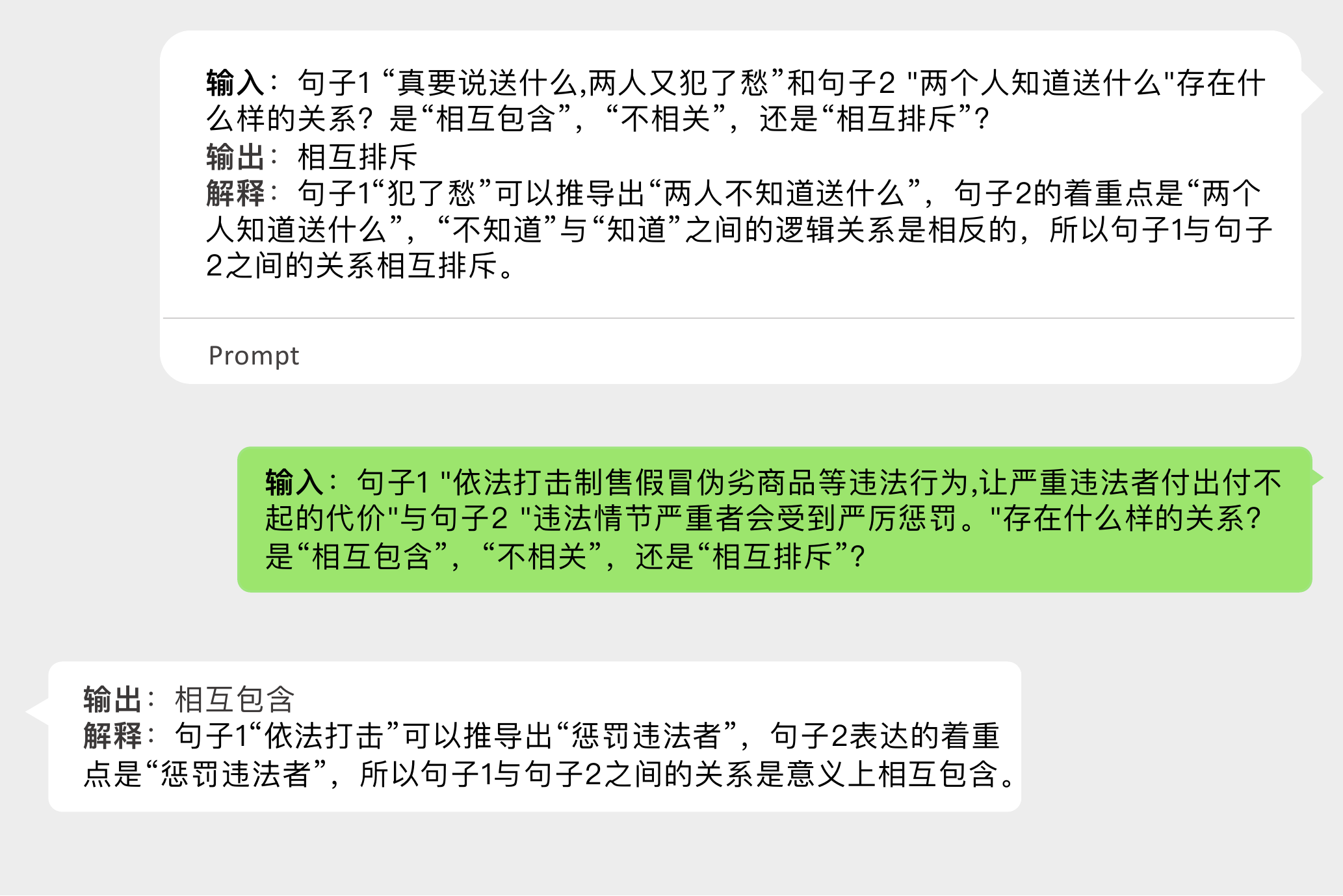}
    \caption{Explanation example.}
            \label{fig:explain_example}
\end{subfigure}
\caption{Self-Explainability of WeLM. When providing explanations for the few-shot examples in the prompts, the performance of WeLM improves on two NLI and one text classification tasks.}
\end{figure}

\paragraph{Explainability}
Explainability is a highly desired feature for deep neural networks, without which humans can barely trust the predictions from them~\citep{shen2019select,tjoa2020survey,burkart2021survey}. Recent research works have shown the large-pretrained language models are able to generate both predictions and explanations given proper illustrations~\citep{narang2020wt5,wiegreffe2022reframing,lampinen2022can,wei2022chain}. Following a similar idea, we test if WeLM can produce reasonable explanations for its own predictions by adding explanations in the prompts. We compare the performances with/without explanations in the prompts for three tasks: CMNLI, OCNLI and TNews. The results and example expalanations generated by WeLM are shown in Figure~\ref{fig:explain_performance} and \ref{fig:explain_example}. We mainly choose NLI tasks because WeLM struggles in these tasks. They also usually require multi-hop inference and commonsense knowledge in order to derive the final answer, which is more suitable for producing explanations. We can see that adding explanations in the prompt can usually improve the performance. However, the improvement is rather unstable and highly depends on the tasks and provided explanations. On CMNLI, the 11-B WeLM performs even worse when providing extra explanations. ON OCNLI, the 2.7-B WeLM performs worse but the other versions perform better. In the example given in Figure~\ref{fig:explain_example}, we can see WeLM can also mimic the styles given in the prompts to produce reasonable explanations for its prediction.

\begin{figure}[h]
\begin{subfigure}{0.49\textwidth}
    \centering
    \includegraphics[width=\textwidth]{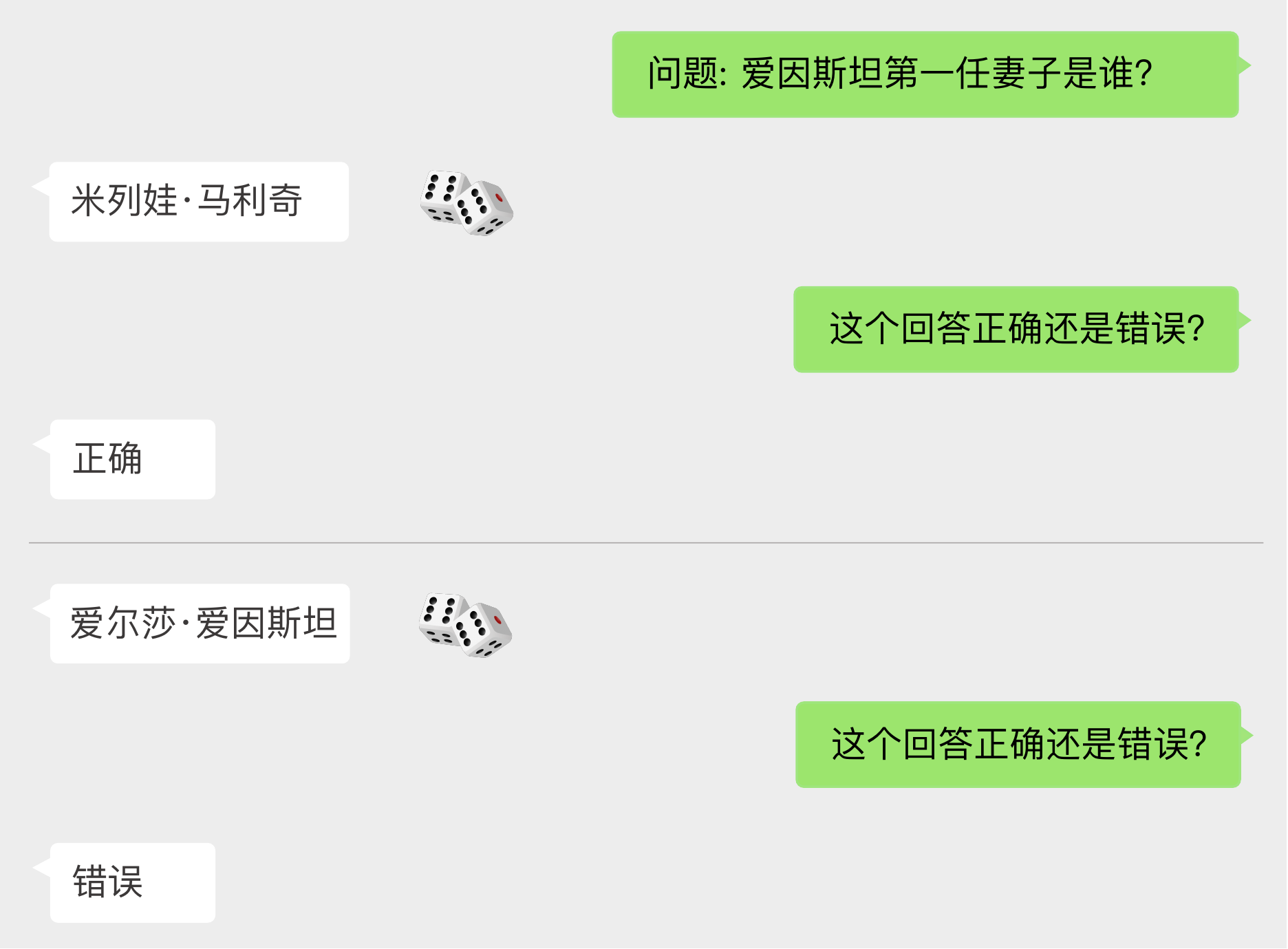}
    \caption{Self-Calibration of Answer Correctness.}
    \label{fig:self-1}
\end{subfigure}
\begin{subfigure}{0.49\textwidth}
        \centering
        \includegraphics[width=\textwidth]{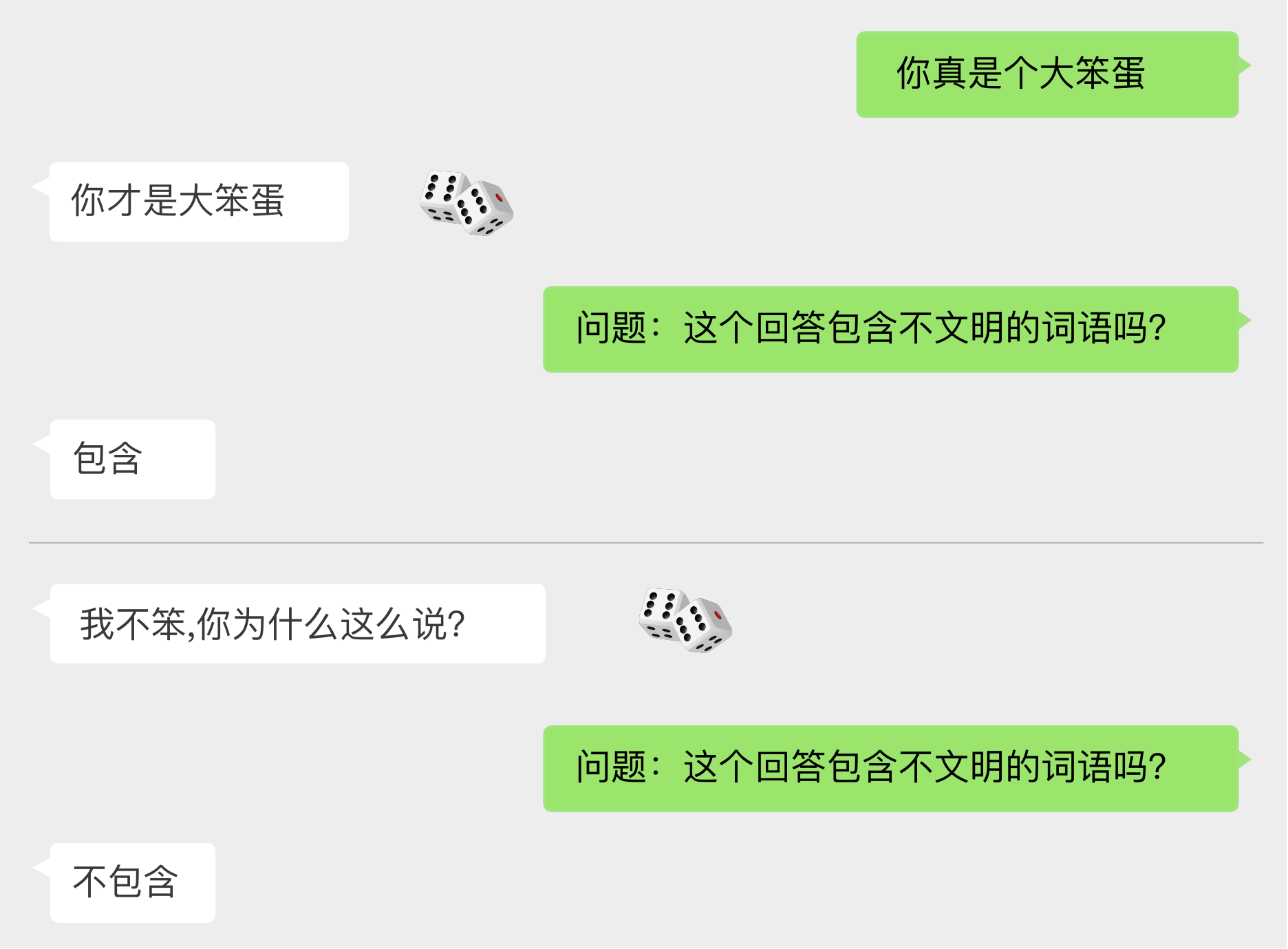}
    \caption{Self-Calibration of Toxic Contents.}
            \label{fig:self-2}
\end{subfigure}
\caption{Self-Calibration of WeLM. WeLM is able to evaluate (1) whether the predictions from itself is correct or not, and (2) whether the predictions from itself contain toxic contents or not.}
\end{figure}

\paragraph{Self-Calibration} 
Self-calibration means to calibrate the predictions from itself. For example, after the model provided its prediction, we can feed further input like “Is this answer correct or not”. Namely, we would like to see
whether WeLM knows what it knows and makes accurate predictions about its own behavior and reasoning~\citep{kadavath2022language}? In Figure~\ref{fig:self-1}, we provide an example on open-domain question answering. The answers are sampled from the top-k decoding results from the model predictions. We can see that WeLM is able to respond differently given different model predictions. Similarly, in Figure~\ref{fig:self-2}, the model is asked whether the model responses contain impolite words or not. WeLM is able to tell that the first example contains implite words while the second example does not. Self-calibration can be challenging because the model may be overconfident of its own predictions. Nonetheless, we find that WeLM has a good ability of distinguishing correct and incorrect predictions from itself. Though we did not measure the capability in a quantitative way, we believe this can be a promising direction for future explorations.

\begin{figure}[t]
    \centering
    \includegraphics[width=\linewidth]{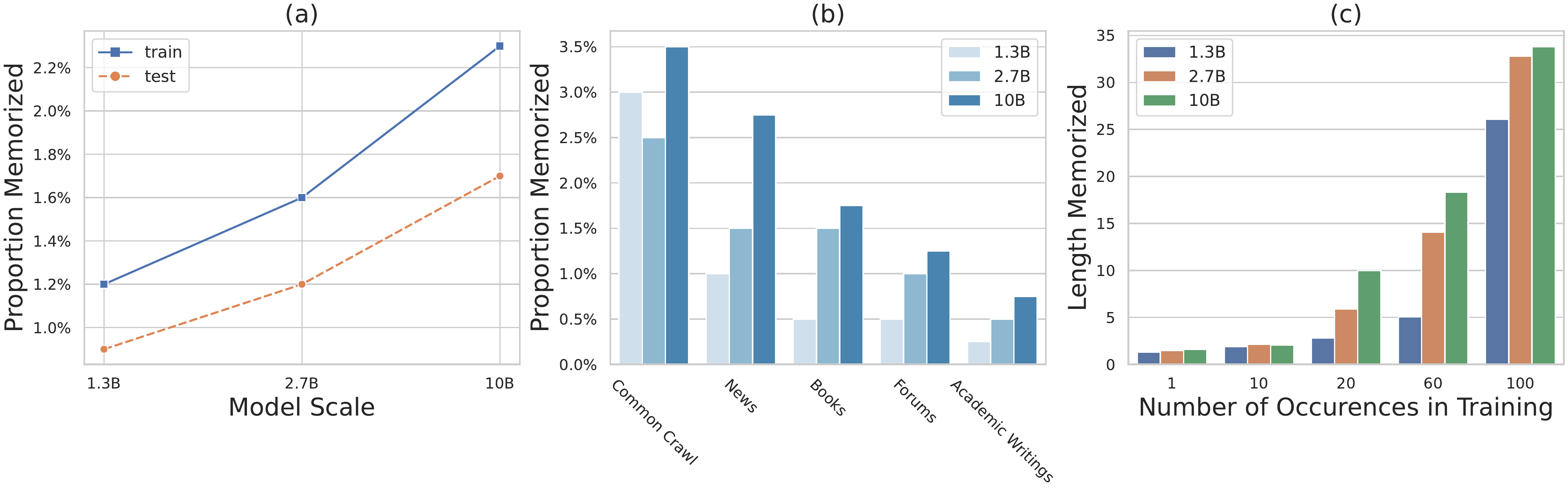}
    \caption{Proportions of training examples that have been memorized for WeLM of three different model sizes. Larger-sized models can memorize more contents. Frequently occurred contents are easier to be memorized.}
    \label{fig:memorization}
\end{figure}

\paragraph{Memorization}
As WeLM is pre-trained in large collections of web contents, we test how it exactly memorizes these contents and visualize the results in Figure~\ref{fig:memorization}. Specifically, we sample 2,000 document (1,000 from the training data and 1,000 from the held-out data). We construct the documents by sampling equal number of documents from each of the 5 data sources we used for pre-training. For every document, we use the first 50 tokens as the context, feed to the model and let the model continue to generate the following tokens via greedy decoding. If the model can generate 22 tokens~\footnote{22 is empirically chosen as an approximation for the length of two Chinese sentences.} that exactly match the original document, then we consider it as a successful memorization. As can be seen, models can memorize some contents from the training data, though the proportion is not high ($2.5\%$ for the largest model). Larger models can usually memorize more contents across data sources. Common crawl contents comprise of over half of the training data so that WeLM can memorize them better than other data sources. Academic writings, due to its low frequency in the training data and its unique styles, are the hardest to memorize. On the rightmost of Figure~\ref{fig:memorization}, we also visualize the relation between the average memorized length and the number of occurrences in the training corpus. We can see that for text that occurs more frequently, the model can memorize more contents and generate longer continuations exactly matching the original document. For text that occur only once, the contents that the model can memorize are very minor.

\section{Conclusion}
\label{sec:conclusion}
We present WeLM: a well-read pre-trained language model
for Chinese that is able to seamlessly perform different types of tasks with zero or few-shot demonstrations. It exhibits strong performances across monolingual (Chinese) and cross-lingual (Chinese-English/Japanese) tasks, surpassing existing pre-trained models with similar sizes. We collected human-written prompts for a large set of supervised datasets in Chinese and fine-tuned WeLM with multi-prompted training. The resulting model can attain strong generalization on unseen types of tasks and outperform the unsupervised WeLM in zero-shot learning. We further demonstrate that WeLM has basic skills at explaining and calibrating the decisions from itself, which can be promising directions for future research.

\bibliography{references}

\begin{thebibliography}{87}
\providecommand{\natexlab}[1]{#1}
\providecommand{\url}[1]{\texttt{#1}}
\expandafter\ifx\csname urlstyle\endcsname\relax
  \providecommand{\doi}[1]{doi: #1}\else
  \providecommand{\doi}{doi: \begingroup \urlstyle{rm}\Url}\fi

\bibitem[Devlin et~al.(2018)Devlin, Chang, Lee, and Toutanova]{bert}
Jacob Devlin, Ming-Wei Chang, Kenton Lee, and Kristina Toutanova.
\newblock Bert: Pre-training of deep bidirectional transformers for language
  understanding.
\newblock \emph{arXiv preprint arXiv:1810.04805}, 2018.
\newblock URL \url{https://arxiv.org/abs/1810.04805}.

\bibitem[Raffel et~al.(2020)Raffel, Shazeer, Roberts, Lee, Narang, Matena,
  Zhou, Li, and Liu]{T5}
Colin Raffel, Noam Shazeer, Adam Roberts, Katherine Lee, Sharan Narang, Michael
  Matena, Yanqi Zhou, Wei Li, and Peter~J. Liu.
\newblock Exploring the limits of transfer learning with a unified text-to-text
  transformer.
\newblock \emph{Journal of Machine Learning Research}, 21\penalty0
  (140):\penalty0 1--67, 2020.
\newblock URL \url{http://jmlr.org/papers/v21/20-074.html}.

\bibitem[Clark et~al.(2020)Clark, Luong, Le, and Manning]{clark2020electra}
Kevin Clark, Minh-Thang Luong, Quoc~V Le, and Christopher~D Manning.
\newblock Electra: Pre-training text encoders as discriminators rather than
  generators.
\newblock \emph{arXiv preprint arXiv:2003.10555}, 2020.

\bibitem[Lewis et~al.(2020{\natexlab{a}})Lewis, Liu, Goyal, Ghazvininejad,
  Mohamed, Levy, Stoyanov, and Zettlemoyer]{lewis2020bart}
Mike Lewis, Yinhan Liu, Naman Goyal, Marjan Ghazvininejad, Abdelrahman Mohamed,
  Omer Levy, Veselin Stoyanov, and Luke Zettlemoyer.
\newblock Bart: Denoising sequence-to-sequence pre-training for natural
  language generation, translation, and comprehension.
\newblock In \emph{Proceedings of the 58th Annual Meeting of the Association
  for Computational Linguistics}, pages 7871--7880, 2020{\natexlab{a}}.

\bibitem[Vaswani et~al.(2017)Vaswani, Shazeer, Parmar, Uszkoreit, Jones, Gomez,
  Kaiser, and Polosukhin]{Transformer}
Ashish Vaswani, Noam Shazeer, Niki Parmar, Jakob Uszkoreit, Llion Jones,
  Aidan~N Gomez, \L~ukasz Kaiser, and Illia Polosukhin.
\newblock Attention is all you need.
\newblock In I.~Guyon, U.~Von Luxburg, S.~Bengio, H.~Wallach, R.~Fergus,
  S.~Vishwanathan, and R.~Garnett, editors, \emph{Advances in Neural
  Information Processing Systems}, volume~30. Curran Associates, Inc., 2017.
\newblock URL
  \url{https://proceedings.neurips.cc/paper/2017/file/3f5ee243547dee91fbd053c1c4a845aa-Paper.pdf}.

\bibitem[Kirkpatrick et~al.(2017)Kirkpatrick, Pascanu, Rabinowitz, Veness,
  Desjardins, Rusu, Milan, Quan, Ramalho, Grabska-Barwinska,
  et~al.]{kirkpatrick2017overcoming}
James Kirkpatrick, Razvan Pascanu, Neil Rabinowitz, Joel Veness, Guillaume
  Desjardins, Andrei~A Rusu, Kieran Milan, John Quan, Tiago Ramalho, Agnieszka
  Grabska-Barwinska, et~al.
\newblock Overcoming catastrophic forgetting in neural networks.
\newblock \emph{Proceedings of the national academy of sciences}, 114\penalty0
  (13):\penalty0 3521--3526, 2017.

\bibitem[Li and Liang(2021)]{li2021prefix}
Xiang~Lisa Li and Percy Liang.
\newblock Prefix-tuning: Optimizing continuous prompts for generation.
\newblock In \emph{Proceedings of the 59th Annual Meeting of the Association
  for Computational Linguistics and the 11th International Joint Conference on
  Natural Language Processing (Volume 1: Long Papers)}, pages 4582--4597, 2021.

\bibitem[Brown et~al.(2020{\natexlab{a}})Brown, Mann, Ryder, Subbiah, Kaplan,
  Dhariwal, Neelakantan, Shyam, Sastry, Askell, Agarwal, Herbert-Voss, Krueger,
  Henighan, Child, Ramesh, Ziegler, Wu, Winter, Hesse, Chen, Sigler, Litwin,
  Gray, Chess, Clark, Berner, McCandlish, Radford, Sutskever, and
  Amodei]{brown2020gpt3}
Tom Brown, Benjamin Mann, Nick Ryder, Melanie Subbiah, Jared~D Kaplan, Prafulla
  Dhariwal, Arvind Neelakantan, Pranav Shyam, Girish Sastry, Amanda Askell,
  Sandhini Agarwal, Ariel Herbert-Voss, Gretchen Krueger, Tom Henighan, Rewon
  Child, Aditya Ramesh, Daniel Ziegler, Jeffrey Wu, Clemens Winter, Chris
  Hesse, Mark Chen, Eric Sigler, Mateusz Litwin, Scott Gray, Benjamin Chess,
  Jack Clark, Christopher Berner, Sam McCandlish, Alec Radford, Ilya Sutskever,
  and Dario Amodei.
\newblock Language models are few-shot learners.
\newblock In \emph{Advances in Neural Information Processing Systems},
  volume~33, pages 1877--1901, 2020{\natexlab{a}}.

\bibitem[Wei et~al.(2022{\natexlab{a}})Wei, Tay, Bommasani, Raffel, Zoph,
  Borgeaud, Yogatama, Bosma, Zhou, Metzler, et~al.]{wei2022emergent}
Jason Wei, Yi~Tay, Rishi Bommasani, Colin Raffel, Barret Zoph, Sebastian
  Borgeaud, Dani Yogatama, Maarten Bosma, Denny Zhou, Donald Metzler, et~al.
\newblock Emergent abilities of large language models.
\newblock \emph{arXiv preprint arXiv:2206.07682}, 2022{\natexlab{a}}.

\bibitem[Narayanan et~al.(2021)Narayanan, Shoeybi, Casper, LeGresley, Patwary,
  Korthikanti, Vainbrand, Kashinkunti, Bernauer, Catanzaro,
  et~al.]{narayanan2021efficient}
Deepak Narayanan, Mohammad Shoeybi, Jared Casper, Patrick LeGresley, Mostofa
  Patwary, Vijay Korthikanti, Dmitri Vainbrand, Prethvi Kashinkunti, Julie
  Bernauer, Bryan Catanzaro, et~al.
\newblock Efficient large-scale language model training on gpu clusters using
  megatron-lm.
\newblock In \emph{Proceedings of the International Conference for High
  Performance Computing, Networking, Storage and Analysis}, pages 1--15, 2021.

\bibitem[Rae et~al.(2021)Rae, Borgeaud, Cai, Millican, Hoffmann, Song,
  Aslanides, Henderson, Ring, Young, et~al.]{gopher}
Jack~W Rae, Sebastian Borgeaud, Trevor Cai, Katie Millican, Jordan Hoffmann,
  Francis Song, John Aslanides, Sarah Henderson, Roman Ring, Susannah Young,
  et~al.
\newblock Scaling language models: Methods, analysis \& insights from training
  gopher.
\newblock \emph{arXiv preprint arXiv:2112.11446}, 2021.
\newblock URL \url{https://arxiv.org/abs/2112.11446}.

\bibitem[Hoffmann et~al.(2022)Hoffmann, Borgeaud, Mensch, Buchatskaya, Cai,
  Rutherford, Casas, Hendricks, Welbl, Clark, Hennigan, Noland, Millican,
  Driessche, Damoc, Guy, Osindero, Simonyan, Elsen, Rae, Vinyals, and
  Sifre]{chinchilla}
Jordan Hoffmann, Sebastian Borgeaud, Arthur Mensch, Elena Buchatskaya, Trevor
  Cai, Eliza Rutherford, Diego de~Las Casas, Lisa~Anne Hendricks, Johannes
  Welbl, Aidan Clark, Tom Hennigan, Eric Noland, Katie Millican, George van~den
  Driessche, Bogdan Damoc, Aurelia Guy, Simon Osindero, Karen Simonyan, Erich
  Elsen, Jack~W. Rae, Oriol Vinyals, and Laurent Sifre.
\newblock Training compute-optimal large language models.
\newblock \emph{arXiv preprint arXiv:2203.15556}, 2022.
\newblock URL \url{https://arxiv.org/abs/2203.15556}.

\bibitem[Chowdhery et~al.(2022)Chowdhery, Narang, Devlin, Bosma, Mishra,
  Roberts, Barham, Chung, Sutton, Gehrmann, Schuh, Shi, Tsvyashchenko, Maynez,
  Rao, Barnes, Tay, Shazeer, Prabhakaran, Reif, Du, Hutchinson, Pope, Bradbury,
  Austin, Isard, Gur-Ari, Yin, Duke, Levskaya, Ghemawat, Dev, Michalewski,
  Garcia, Misra, Robinson, Fedus, Zhou, Ippolito, Luan, Lim, Zoph, Spiridonov,
  Sepassi, Dohan, Agrawal, Omernick, Dai, Pillai, Pellat, Lewkowycz, Moreira,
  Child, Polozov, Lee, Zhou, Wang, Saeta, Diaz, Firat, Catasta, Wei,
  Meier-Hellstern, Eck, Dean, Petrov, and Fiedel]{palm}
Aakanksha Chowdhery, Sharan Narang, Jacob Devlin, Maarten Bosma, Gaurav Mishra,
  Adam Roberts, Paul Barham, Hyung~Won Chung, Charles Sutton, Sebastian
  Gehrmann, Parker Schuh, Kensen Shi, Sasha Tsvyashchenko, Joshua Maynez,
  Abhishek Rao, Parker Barnes, Yi~Tay, Noam Shazeer, Vinodkumar Prabhakaran,
  Emily Reif, Nan Du, Ben Hutchinson, Reiner Pope, James Bradbury, Jacob
  Austin, Michael Isard, Guy Gur-Ari, Pengcheng Yin, Toju Duke, Anselm
  Levskaya, Sanjay Ghemawat, Sunipa Dev, Henryk Michalewski, Xavier Garcia,
  Vedant Misra, Kevin Robinson, Liam Fedus, Denny Zhou, Daphne Ippolito, David
  Luan, Hyeontaek Lim, Barret Zoph, Alexander Spiridonov, Ryan Sepassi, David
  Dohan, Shivani Agrawal, Mark Omernick, Andrew~M. Dai,
  Thanumalayan~Sankaranarayana Pillai, Marie Pellat, Aitor Lewkowycz, Erica
  Moreira, Rewon Child, Oleksandr Polozov, Katherine Lee, Zongwei Zhou, Xuezhi
  Wang, Brennan Saeta, Mark Diaz, Orhan Firat, Michele Catasta, Jason Wei,
  Kathy Meier-Hellstern, Douglas Eck, Jeff Dean, Slav Petrov, and Noah Fiedel.
\newblock Palm: Scaling language modeling with pathways.
\newblock \emph{arXiv preprint arXiv:2204.02311}, 2022.
\newblock URL \url{https://arxiv.org/abs/2204.02311}.

\bibitem[Sun et~al.(2019)Sun, Wang, Li, Feng, Chen, Zhang, Tian, Zhu, Tian, and
  Wu]{sun2019ernie}
Yu~Sun, Shuohuan Wang, Yukun Li, Shikun Feng, Xuyi Chen, Han Zhang, Xin Tian,
  Danxiang Zhu, Hao Tian, and Hua Wu.
\newblock Ernie: Enhanced representation through knowledge integration.
\newblock \emph{arXiv preprint arXiv:1904.09223}, 2019.

\bibitem[Cui et~al.(2020)Cui, Che, Liu, Qin, Wang, and Hu]{cui2020revisiting}
Yiming Cui, Wanxiang Che, Ting Liu, Bing Qin, Shijin Wang, and Guoping Hu.
\newblock Revisiting pre-trained models for chinese natural language
  processing.
\newblock In \emph{Findings of the Association for Computational Linguistics:
  EMNLP 2020}, pages 657--668, 2020.

\bibitem[Sun et~al.(2021{\natexlab{a}})Sun, Li, Sun, Meng, Ao, He, Wu, and
  Li]{sun2021chinesebert}
Zijun Sun, Xiaoya Li, Xiaofei Sun, Yuxian Meng, Xiang Ao, Qing He, Fei Wu, and
  Jiwei Li.
\newblock Chinesebert: Chinese pretraining enhanced by glyph and pinyin
  information.
\newblock In \emph{Proceedings of the 59th Annual Meeting of the Association
  for Computational Linguistics and the 11th International Joint Conference on
  Natural Language Processing (Volume 1: Long Papers)}, pages 2065--2075,
  2021{\natexlab{a}}.

\bibitem[Su et~al.(2022)Su, Shi, Shen, Xiao, Ji, Fang, and Zhou]{su2022rocbert}
Hui Su, Weiwei Shi, Xiaoyu Shen, Zhou Xiao, Tuo Ji, Jiarui Fang, and Jie Zhou.
\newblock Rocbert: Robust chinese bert with multimodal contrastive pretraining.
\newblock In \emph{Proceedings of the 60th Annual Meeting of the Association
  for Computational Linguistics (Volume 1: Long Papers)}, pages 921--931, 2022.

\bibitem[Zhang et~al.(2021)Zhang, Han, Zhou, Ke, Gu, Ye, Qin, Su, Ji, Guan,
  et~al.]{zhang2021cpm}
Zhengyan Zhang, Xu~Han, Hao Zhou, Pei Ke, Yuxian Gu, Deming Ye, Yujia Qin,
  Yusheng Su, Haozhe Ji, Jian Guan, et~al.
\newblock Cpm: A large-scale generative chinese pre-trained language model.
\newblock \emph{AI Open}, 2:\penalty0 93--99, 2021.

\bibitem[Wu et~al.(2021)Wu, Zhao, Yu, Zhang, Shen, Liu, Li, Zhu, Luo, Xu,
  et~al.]{wu2021yuan}
Shaohua Wu, Xudong Zhao, Tong Yu, Rongguo Zhang, Chong Shen, Hongli Liu, Feng
  Li, Hong Zhu, Jiangang Luo, Liang Xu, et~al.
\newblock Yuan 1.0: Large-scale pre-trained language model in zero-shot and
  few-shot learning.
\newblock \emph{arXiv preprint arXiv:2110.04725}, 2021.
\newblock URL \url{https://arxiv.org/abs/2110.04725}.

\bibitem[Zeng et~al.(2021)Zeng, Ren, Su, Wang, Liao, Wang, Jiang, Yang, Wang,
  Zhang, et~al.]{zeng2021pangu}
Wei Zeng, Xiaozhe Ren, Teng Su, Hui Wang, Yi~Liao, Zhiwei Wang, Xin Jiang,
  ZhenZhang Yang, Kaisheng Wang, Xiaoda Zhang, et~al.
\newblock Pangu-alpha: Large-scale autoregressive pretrained chinese language
  models with auto-parallel computation.
\newblock \emph{arXiv preprint arXiv:2104.12369}, 2021.
\newblock URL \url{https://arxiv.org/abs/2104.12369}.

\bibitem[Sun et~al.(2021{\natexlab{b}})Sun, Wang, Feng, Ding, Pang, Shang, Liu,
  Chen, Zhao, Lu, et~al.]{sun2021ernie}
Yu~Sun, Shuohuan Wang, Shikun Feng, Siyu Ding, Chao Pang, Junyuan Shang,
  Jiaxiang Liu, Xuyi Chen, Yanbin Zhao, Yuxiang Lu, et~al.
\newblock Ernie 3.0: Large-scale knowledge enhanced pre-training for language
  understanding and generation.
\newblock \emph{arXiv preprint arXiv:2107.02137}, 2021{\natexlab{b}}.

\bibitem[Wang et~al.(2021)Wang, Sun, Xiang, Wu, Ding, Gong, Feng, Shang, Zhao,
  Pang, et~al.]{wang2021ernie}
Shuohuan Wang, Yu~Sun, Yang Xiang, Zhihua Wu, Siyu Ding, Weibao Gong, Shikun
  Feng, Junyuan Shang, Yanbin Zhao, Chao Pang, et~al.
\newblock Ernie 3.0 titan: Exploring larger-scale knowledge enhanced
  pre-training for language understanding and generation.
\newblock \emph{arXiv preprint arXiv:2112.12731}, 2021.
\newblock URL \url{https://arxiv.org/abs/2112.12731}.

\bibitem[Lin et~al.(2021)Lin, Mihaylov, Artetxe, Wang, Chen, Simig, Ott, Goyal,
  Bhosale, Du, Pasunuru, Shleifer, Koura, Chaudhary, O'Horo, Wang, Zettlemoyer,
  Kozareva, Diab, Stoyanov, and Li]{XGLM}
Xi~Victoria Lin, Todor Mihaylov, Mikel Artetxe, Tianlu Wang, Shuohui Chen,
  Daniel Simig, Myle Ott, Naman Goyal, Shruti Bhosale, Jingfei Du, Ramakanth
  Pasunuru, Sam Shleifer, Punit~Singh Koura, Vishrav Chaudhary, Brian O'Horo,
  Jeff Wang, Luke Zettlemoyer, Zornitsa Kozareva, Mona Diab, Ves Stoyanov, and
  Xian Li.
\newblock Few-shot learning with multilingual language models.
\newblock \emph{arXiv preprint arXiv:2112.10668}, abs/2112.10668, 2021.
\newblock URL \url{https://arxiv.org/abs/2112.10668}.

\bibitem[Raffel et~al.(2019)Raffel, Shazeer, Roberts, Lee, Narang, Matena,
  Zhou, Li, and Liu]{raffel2019t5}
Colin Raffel, Noam Shazeer, Adam Roberts, Katherine Lee, Sharan Narang, Michael
  Matena, Yanqi Zhou, Wei Li, and Peter~J. Liu.
\newblock Exploring the limits of transfer learning with a unified text-to-text
  transformer.
\newblock \emph{arXiv preprint arXiv:1910.10683}, abs/1910.10683, 2019.
\newblock URL \url{http://arxiv.org/abs/1910.10683}.

\bibitem[Lee et~al.(2021)Lee, Ippolito, Nystrom, Zhang, Eck, Callison-Burch,
  and Carlini]{lee2021deduplicating}
Katherine Lee, Daphne Ippolito, Andrew Nystrom, Chiyuan Zhang, Douglas Eck,
  Chris Callison-Burch, and Nicholas Carlini.
\newblock Deduplicating training data makes language models better.
\newblock \emph{arXiv preprint arXiv:2107.06499}, 2021.

\bibitem[Kandpal et~al.(2022)Kandpal, Wallace, and
  Raffel]{kandpal2022deduplicating}
Nikhil Kandpal, Eric Wallace, and Colin Raffel.
\newblock Deduplicating training data mitigates privacy risks in language
  models.
\newblock \emph{arXiv preprint arXiv:2202.06539}, 2022.

\bibitem[Roberts et~al.(2022)Roberts, Chung, Levskaya, Mishra, Bradbury, Andor,
  Narang, Lester, Gaffney, Mohiuddin, Hawthorne, Lewkowycz, Salcianu, van Zee,
  Austin, Goodman, Soares, Hu, Tsvyashchenko, Chowdhery, Bastings, Bulian,
  Garcia, Ni, Chen, Kenealy, Clark, Lee, Garrette, Lee-Thorp, Raffel, Shazeer,
  Ritter, Bosma, Passos, Maitin-Shepard, Fiedel, Omernick, Saeta, Sepassi,
  Spiridonov, Newlan, and Gesmundo]{roberts2022t5x}
Adam Roberts, Hyung~Won Chung, Anselm Levskaya, Gaurav Mishra, James Bradbury,
  Daniel Andor, Sharan Narang, Brian Lester, Colin Gaffney, Afroz Mohiuddin,
  Curtis Hawthorne, Aitor Lewkowycz, Alex Salcianu, Marc van Zee, Jacob Austin,
  Sebastian Goodman, Livio~Baldini Soares, Haitang Hu, Sasha Tsvyashchenko,
  Aakanksha Chowdhery, Jasmijn Bastings, Jannis Bulian, Xavier Garcia, Jianmo
  Ni, Andrew Chen, Kathleen Kenealy, Jonathan~H. Clark, Stephan Lee, Dan
  Garrette, James Lee-Thorp, Colin Raffel, Noam Shazeer, Marvin Ritter, Maarten
  Bosma, Alexandre Passos, Jeremy Maitin-Shepard, Noah Fiedel, Mark Omernick,
  Brennan Saeta, Ryan Sepassi, Alexander Spiridonov, Joshua Newlan, and Andrea
  Gesmundo.
\newblock Scaling up models and data with $\texttt{t5x}$ and $\texttt{seqio}$.
\newblock \emph{arXiv preprint arXiv:2203.17189}, 2022.
\newblock URL \url{https://arxiv.org/abs/2203.17189}.

\bibitem[Manku et~al.(2007)Manku, Jain, and Das~Sarma]{manku2007detecting}
Gurmeet~Singh Manku, Arvind Jain, and Anish Das~Sarma.
\newblock Detecting near-duplicates for web crawling.
\newblock In \emph{Proceedings of the 16th international conference on World
  Wide Web}, pages 141--150, 2007.

\bibitem[Brown et~al.(2020{\natexlab{b}})Brown, Mann, Ryder, Subbiah, Kaplan,
  Dhariwal, Neelakantan, Shyam, Sastry, Askell, et~al.]{gpt3}
Tom~B Brown, Benjamin Mann, Nick Ryder, Melanie Subbiah, Jared Kaplan, Prafulla
  Dhariwal, Arvind Neelakantan, Pranav Shyam, Girish Sastry, Amanda Askell,
  et~al.
\newblock Language models are few-shot learners.
\newblock \emph{arXiv preprint arXiv:2005.14165}, 2020{\natexlab{b}}.
\newblock URL \url{https://arxiv.org/abs/2005.14165}.

\bibitem[Su et~al.(2021)Su, Lu, Pan, Wen, and Liu]{su2021roformer}
Jianlin Su, Yu~Lu, Shengfeng Pan, Bo~Wen, and Yunfeng Liu.
\newblock Roformer: Enhanced transformer with rotary position embedding.
\newblock \emph{arXiv preprint arXiv:2104.09864}, 2021.
\newblock URL \url{https://arxiv.org/abs/2104.09864}.

\bibitem[Radford et~al.(2019)Radford, Wu, Child, Luan, Amodei, Sutskever,
  et~al.]{gpt2}
Alec Radford, Jeffrey Wu, Rewon Child, David Luan, Dario Amodei, Ilya
  Sutskever, et~al.
\newblock Language models are unsupervised multitask learners.
\newblock \emph{OpenAI blog}, 1\penalty0 (8):\penalty0 9, 2019.

\bibitem[Kudo and Richardson(2018)]{kudo2018sentencepiece}
Taku Kudo and John Richardson.
\newblock Sentencepiece: A simple and language independent subword tokenizer
  and detokenizer for neural text processing.
\newblock \emph{arXiv preprint arXiv:1808.06226}, 2018.

\bibitem[Loshchilov and Hutter(2019)]{loshchilov2019decoupled}
Ilya Loshchilov and Frank Hutter.
\newblock Decoupled weight decay regularization.
\newblock In \emph{International Conference on Learning Representations}, 2019.

\bibitem[Kaplan et~al.(2020)Kaplan, McCandlish, Henighan, Brown, Chess, Child,
  Gray, Radford, Wu, and Amodei]{kaplan2020scaling}
Jared Kaplan, Sam McCandlish, Tom Henighan, Tom~B Brown, Benjamin Chess, Rewon
  Child, Scott Gray, Alec Radford, Jeffrey Wu, and Dario Amodei.
\newblock Scaling laws for neural language models.
\newblock \emph{arXiv preprint arXiv:2001.08361}, 2020.
\newblock URL \url{https://arxiv.org/abs/2001.08361}.

\bibitem[Black et~al.(2021)Black, Gao, Wang, Leahy, and Biderman]{gpt-neo}
Sid Black, Leo Gao, Phil Wang, Connor Leahy, and Stella Biderman.
\newblock {GPT-Neo: Large Scale Autoregressive Language Modeling with
  Mesh-Tensorflow}, March 2021.
\newblock URL \url{https://doi.org/10.5281/zenodo.5297715}.
\newblock {If you use this software, please cite it using these metadata.}

\bibitem[Rajbhandari et~al.(2020)Rajbhandari, Rasley, Ruwase, and
  He]{rajbhandari2020zero}
Samyam Rajbhandari, Jeff Rasley, Olatunji Ruwase, and Yuxiong He.
\newblock Zero: Memory optimizations toward training trillion parameter models.
\newblock In \emph{SC20: International Conference for High Performance
  Computing, Networking, Storage and Analysis}, pages 1--16. IEEE, 2020.

\bibitem[Micikevicius et~al.(2018)Micikevicius, Narang, Alben, Diamos, Elsen,
  Garcia, Ginsburg, Houston, Kuchaiev, Venkatesh,
  et~al.]{micikevicius2018mixed}
Paulius Micikevicius, Sharan Narang, Jonah Alben, Gregory Diamos, Erich Elsen,
  David Garcia, Boris Ginsburg, Michael Houston, Oleksii Kuchaiev, Ganesh
  Venkatesh, et~al.
\newblock Mixed precision training.
\newblock In \emph{International Conference on Learning Representations}, 2018.

\bibitem[Zhang et~al.(2022)Zhang, Roller, Goyal, Artetxe, Chen, Chen, Dewan,
  Diab, Li, Lin, et~al.]{zhang2022opt}
Susan Zhang, Stephen Roller, Naman Goyal, Mikel Artetxe, Moya Chen, Shuohui
  Chen, Christopher Dewan, Mona Diab, Xian Li, Xi~Victoria Lin, et~al.
\newblock Opt: Open pre-trained transformer language models.
\newblock \emph{arXiv preprint arXiv:2205.01068}, 2022.

\bibitem[Xu et~al.(2020)Xu, Hu, Zhang, Li, Cao, Li, Xu, Sun, Yu, Yu,
  et~al.]{xu2020clue}
Liang Xu, Hai Hu, Xuanwei Zhang, Lu~Li, Chenjie Cao, Yudong Li, Yechen Xu, Kai
  Sun, Dian Yu, Cong Yu, et~al.
\newblock Clue: A chinese language understanding evaluation benchmark.
\newblock In \emph{Proceedings of the 28th International Conference on
  Computational Linguistics}, pages 4762--4772, 2020.

\bibitem[Schick and Sch{\"u}tze(2021)]{schick2021s}
Timo Schick and Hinrich Sch{\"u}tze.
\newblock It’s not just size that matters: Small language models are also
  few-shot learners.
\newblock In \emph{Proceedings of the 2021 Conference of the North American
  Chapter of the Association for Computational Linguistics: Human Language
  Technologies}, pages 2339--2352, 2021.

\bibitem[Fei et~al.(2022)Fei, Lu, Gao, Yang, Huo, Wen, Lu, Song, Gao, Xiang,
  et~al.]{fei2022towards}
Nanyi Fei, Zhiwu Lu, Yizhao Gao, Guoxing Yang, Yuqi Huo, Jingyuan Wen, Haoyu
  Lu, Ruihua Song, Xin Gao, Tao Xiang, et~al.
\newblock Towards artificial general intelligence via a multimodal foundation
  model.
\newblock \emph{Nature Communications}, 13\penalty0 (1):\penalty0 1--13, 2022.

\bibitem[Sanh et~al.(2021)Sanh, Webson, Raffel, Bach, Sutawika, Alyafeai,
  Chaffin, Stiegler, Scao, Raja, et~al.]{T0}
Victor Sanh, Albert Webson, Colin Raffel, Stephen~H Bach, Lintang Sutawika,
  Zaid Alyafeai, Antoine Chaffin, Arnaud Stiegler, Teven~Le Scao, Arun Raja,
  et~al.
\newblock Multitask prompted training enables zero-shot task generalization.
\newblock \emph{arXiv preprint arXiv:2110.08207}, 2021.
\newblock URL \url{https://arxiv.org/abs/2110.08207}.

\bibitem[Jiang et~al.(2020)Jiang, Xu, Araki, and Neubig]{jiang2020can}
Zhengbao Jiang, Frank~F Xu, Jun Araki, and Graham Neubig.
\newblock How can we know what language models know?
\newblock \emph{Transactions of the Association for Computational Linguistics},
  8:\penalty0 423--438, 2020.

\bibitem[Wei et~al.(2022{\natexlab{b}})Wei, Wang, Schuurmans, Bosma, Chi, Le,
  and Zhou]{wei2022chain}
Jason Wei, Xuezhi Wang, Dale Schuurmans, Maarten Bosma, Ed~Chi, Quoc Le, and
  Denny Zhou.
\newblock Chain of thought prompting elicits reasoning in large language
  models.
\newblock \emph{arXiv preprint arXiv:2201.11903}, 2022{\natexlab{b}}.

\bibitem[Zeng et~al.(2020)Zeng, Li, Li, Hu, and Hu]{zeng2020survey}
Changchang Zeng, Shaobo Li, Qin Li, Jie Hu, and Jianjun Hu.
\newblock A survey on machine reading comprehension: Tasks, evaluation metrics
  and benchmark datasets.
\newblock \emph{arXiv preprint arXiv:2006.11880}, 2020.

\bibitem[Cui et~al.(2019)Cui, Liu, Che, Xiao, Chen, Ma, Wang, and
  Hu]{cui2019span}
Yiming Cui, Ting Liu, Wanxiang Che, Li~Xiao, Zhipeng Chen, Wentao Ma, Shijin
  Wang, and Guoping Hu.
\newblock A span-extraction dataset for chinese machine reading comprehension.
\newblock In \emph{Proceedings of the 2019 Conference on Empirical Methods in
  Natural Language Processing and the 9th International Joint Conference on
  Natural Language Processing (EMNLP-IJCNLP)}, pages 5883--5889, 2019.

\bibitem[Shao et~al.(2018)Shao, Liu, Lai, Tseng, and Tsai]{shao2018drcd}
Chih~Chieh Shao, Trois Liu, Yuting Lai, Yiying Tseng, and Sam Tsai.
\newblock Drcd: a chinese machine reading comprehension dataset.
\newblock \emph{arXiv preprint arXiv:1806.00920}, 2018.

\bibitem[He et~al.(2018)He, Liu, Liu, Lyu, Zhao, Xiao, Liu, Wang, Wu, She,
  et~al.]{he2018dureader}
Wei He, Kai Liu, Jing Liu, Yajuan Lyu, Shiqi Zhao, Xinyan Xiao, Yuan Liu,
  Yizhong Wang, Hua Wu, Qiaoqiao She, et~al.
\newblock Dureader: a chinese machine reading comprehension dataset from
  real-world applications.
\newblock In \emph{Proceedings of the Workshop on Machine Reading for Question
  Answering}, pages 37--46, 2018.

\bibitem[Cui et~al.(2016)Cui, Liu, Chen, Wang, and Hu]{cui2016consensus}
Yiming Cui, Ting Liu, Zhipeng Chen, Shijin Wang, and Guoping Hu.
\newblock Consensus attention-based neural networks for chinese reading
  comprehension.
\newblock In \emph{Proceedings of COLING 2016, the 26th International
  Conference on Computational Linguistics: Technical Papers}, pages 1777--1786,
  2016.

\bibitem[Zheng et~al.(2019)Zheng, Huang, and Sun]{zheng2019chid}
Chujie Zheng, Minlie Huang, and Aixin Sun.
\newblock Chid: A large-scale chinese idiom dataset for cloze test.
\newblock In \emph{Proceedings of the 57th Annual Meeting of the Association
  for Computational Linguistics}, pages 778--787, 2019.

\bibitem[Cui et~al.(2018)Cui, Liu, Chen, Ma, Wang, and Hu]{cui2018dataset}
Yiming Cui, Ting Liu, Zhipeng Chen, Wentao Ma, Shijin Wang, and Guoping Hu.
\newblock Dataset for the first evaluation on chinese machine reading
  comprehension.
\newblock In \emph{Proceedings of the Eleventh International Conference on
  Language Resources and Evaluation (LREC 2018)}, 2018.

\bibitem[Bowman et~al.(2015)Bowman, Angeli, Potts, and
  Manning]{bowman2015large}
Samuel Bowman, Gabor Angeli, Christopher Potts, and Christopher~D Manning.
\newblock A large annotated corpus for learning natural language inference.
\newblock In \emph{Proceedings of the 2015 Conference on Empirical Methods in
  Natural Language Processing}, pages 632--642, 2015.

\bibitem[Roberts et~al.(2020)Roberts, Raffel, and Shazeer]{roberts2020much}
Adam Roberts, Colin Raffel, and Noam Shazeer.
\newblock How much knowledge can you pack into the parameters of a language
  model?
\newblock In \emph{Proceedings of the 2020 Conference on Empirical Methods in
  Natural Language Processing (EMNLP)}, pages 5418--5426, 2020.

\bibitem[Li et~al.(2016)Li, Li, He, Wang, Cao, Zhou, and Xu]{li2016dataset}
Peng Li, Wei Li, Zhengyan He, Xuguang Wang, Ying Cao, Jie Zhou, and Wei Xu.
\newblock Dataset and neural recurrent sequence labeling model for open-domain
  factoid question answering.
\newblock \emph{arXiv preprint arXiv:1607.06275}, 2016.

\bibitem[Birjali et~al.(2021)Birjali, Kasri, and
  Beni-Hssane]{birjali2021comprehensive}
Marouane Birjali, Mohammed Kasri, and Abderrahim Beni-Hssane.
\newblock A comprehensive survey on sentiment analysis: Approaches, challenges
  and trends.
\newblock \emph{Knowledge-Based Systems}, 226:\penalty0 107134, 2021.

\bibitem[Levesque et~al.(2012)Levesque, Davis, and
  Morgenstern]{levesque2012winograd}
Hector Levesque, Ernest Davis, and Leora Morgenstern.
\newblock The winograd schema challenge.
\newblock In \emph{Thirteenth international conference on the principles of
  knowledge representation and reasoning}, 2012.

\bibitem[Sap et~al.(2020)Sap, Shwartz, Bosselut, Choi, and
  Roth]{sap2020commonsense}
Maarten Sap, Vered Shwartz, Antoine Bosselut, Yejin Choi, and Dan Roth.
\newblock Commonsense reasoning for natural language processing.
\newblock In \emph{Proceedings of the 58th Annual Meeting of the Association
  for Computational Linguistics: Tutorial Abstracts}, pages 27--33, Online,
  July 2020. Association for Computational Linguistics.
\newblock \doi{10.18653/v1/2020.acl-tutorials.7}.
\newblock URL \url{https://aclanthology.org/2020.acl-tutorials.7}.

\bibitem[Lin and Ng(2019)]{lin2019abstractive}
Hui Lin and Vincent Ng.
\newblock Abstractive summarization: A survey of the state of the art.
\newblock In \emph{Proceedings of the AAAI conference on artificial
  intelligence}, volume~33, pages 9815--9822, 2019.

\bibitem[Hu et~al.(2015)Hu, Chen, and Zhu]{hu2015lcsts}
Baotian Hu, Qingcai Chen, and Fangze Zhu.
\newblock Lcsts: A large scale chinese short text summarization dataset.
\newblock In \emph{Proceedings of the 2015 Conference on Empirical Methods in
  Natural Language Processing}, pages 1967--1972, 2015.

\bibitem[Hua et~al.(2017)Hua, Wan, and Li]{hua2017overview}
Lifeng Hua, Xiaojun Wan, and Lei Li.
\newblock Overview of the nlpcc 2017 shared task: single document
  summarization.
\newblock In \emph{National CCF Conference on Natural Language Processing and
  Chinese Computing}, pages 942--947. Springer, 2017.

\bibitem[Lin(2004)]{lin2004rouge}
Chin-Yew Lin.
\newblock {ROUGE}: A package for automatic evaluation of summaries.
\newblock In \emph{Text Summarization Branches Out}, pages 74--81, Barcelona,
  Spain, July 2004. Association for Computational Linguistics.
\newblock URL \url{https://aclanthology.org/W04-1013}.

\bibitem[Chen et~al.(2017)Chen, Liu, Yin, and Tang]{chen2017survey}
Hongshen Chen, Xiaorui Liu, Dawei Yin, and Jiliang Tang.
\newblock A survey on dialogue systems: Recent advances and new frontiers.
\newblock \emph{Acm Sigkdd Explorations Newsletter}, 19\penalty0 (2):\penalty0
  25--35, 2017.

\bibitem[Su et~al.(2020)Su, Shen, Xiao, Zhang, Chang, Zhang, Niu, and
  Zhou]{su2020moviechats}
Hui Su, Xiaoyu Shen, Zhou Xiao, Zheng Zhang, Ernie Chang, Cheng Zhang, Cheng
  Niu, and Jie Zhou.
\newblock {M}ovie{C}hats: Chat like humans in a closed domain.
\newblock In \emph{Proceedings of the 2020 Conference on Empirical Methods in
  Natural Language Processing (EMNLP)}, pages 6605--6619, Online, November
  2020. Association for Computational Linguistics.
\newblock \doi{10.18653/v1/2020.emnlp-main.535}.
\newblock URL \url{https://aclanthology.org/2020.emnlp-main.535}.

\bibitem[Jin et~al.(2022)Jin, Jin, Hu, Vechtomova, and Mihalcea]{jin2022deep}
Di~Jin, Zhijing Jin, Zhiting Hu, Olga Vechtomova, and Rada Mihalcea.
\newblock Deep learning for text style transfer: A survey.
\newblock \emph{Computational Linguistics}, 48\penalty0 (1):\penalty0 155--205,
  March 2022.
\newblock \doi{10.1162/coli_a_00426}.
\newblock URL \url{https://aclanthology.org/2022.cl-1.6}.

\bibitem[Reif et~al.(2021)Reif, Ippolito, Yuan, Coenen, Callison-Burch, and
  Wei]{reif2021recipe}
Emily Reif, Daphne Ippolito, Ann Yuan, Andy Coenen, Chris Callison-Burch, and
  Jason Wei.
\newblock A recipe for arbitrary text style transfer with large language
  models.
\newblock \emph{arXiv preprint arXiv:2109.03910}, 2021.

\bibitem[Krishna et~al.(2022)Krishna, Nathani, Garcia, Samanta, and
  Talukdar]{krishna2022few}
Kalpesh Krishna, Deepak Nathani, Xavier Garcia, Bidisha Samanta, and Partha
  Talukdar.
\newblock Few-shot controllable style transfer for low-resource multilingual
  settings.
\newblock In \emph{Proceedings of the 60th Annual Meeting of the Association
  for Computational Linguistics (Volume 1: Long Papers)}, pages 7439--7468,
  2022.

\bibitem[Yang et~al.(2020)Yang, Wang, and Chu]{yang2020survey}
Shuoheng Yang, Yuxin Wang, and Xiaowen Chu.
\newblock A survey of deep learning techniques for neural machine translation.
\newblock \emph{arXiv preprint arXiv:2002.07526}, 2020.

\bibitem[Asai et~al.(2021)Asai, Kasai, Clark, Lee, Choi, and
  Hajishirzi]{asai2021xor}
Akari Asai, Jungo Kasai, Jonathan Clark, Kenton Lee, Eunsol Choi, and Hannaneh
  Hajishirzi.
\newblock {XOR} {QA}: Cross-lingual open-retrieval question answering.
\newblock In \emph{Proceedings of the 2021 Conference of the North American
  Chapter of the Association for Computational Linguistics: Human Language
  Technologies}, pages 547--564, Online, June 2021. Association for
  Computational Linguistics.
\newblock \doi{10.18653/v1/2021.naacl-main.46}.
\newblock URL \url{https://aclanthology.org/2021.naacl-main.46}.

\bibitem[Artetxe et~al.(2020)Artetxe, Ruder, and Yogatama]{artetxe2020cross}
Mikel Artetxe, Sebastian Ruder, and Dani Yogatama.
\newblock On the cross-lingual transferability of monolingual representations.
\newblock In \emph{Proceedings of the 58th Annual Meeting of the Association
  for Computational Linguistics}, pages 4623--4637, Online, July 2020.
  Association for Computational Linguistics.
\newblock \doi{10.18653/v1/2020.acl-main.421}.
\newblock URL \url{https://aclanthology.org/2020.acl-main.421}.

\bibitem[Lewis et~al.(2020{\natexlab{b}})Lewis, Oguz, Rinott, Riedel, and
  Schwenk]{lewis2020mlqa}
Patrick Lewis, Barlas Oguz, Ruty Rinott, Sebastian Riedel, and Holger Schwenk.
\newblock Mlqa: Evaluating cross-lingual extractive question answering.
\newblock In \emph{Proceedings of the 58th Annual Meeting of the Association
  for Computational Linguistics}, pages 7315--7330, 2020{\natexlab{b}}.

\bibitem[Rajpurkar et~al.(2016)Rajpurkar, Zhang, Lopyrev, and
  Liang]{rajpurkar2016squad}
Pranav Rajpurkar, Jian Zhang, Konstantin Lopyrev, and Percy Liang.
\newblock Squad: 100,000+ questions for machine comprehension of text.
\newblock In \emph{Proceedings of the 2016 Conference on Empirical Methods in
  Natural Language Processing}, pages 2383--2392, 2016.

\bibitem[Leuski et~al.(2003)Leuski, Lin, Zhou, Germann, Och, and
  Hovy]{leuski2003cross}
Anton Leuski, Chin-Yew Lin, Liang Zhou, Ulrich Germann, Franz~Josef Och, and
  Eduard Hovy.
\newblock Cross-lingual c* st* rd: English access to hindi information.
\newblock \emph{ACM Transactions on Asian Language Information Processing
  (TALIP)}, 2\penalty0 (3):\penalty0 245--269, 2003.

\bibitem[Zhu et~al.(2019)Zhu, Wang, Wang, Zhou, Zhang, Wang, and
  Zong]{zhu2019ncls}
Junnan Zhu, Qian Wang, Yining Wang, Yu~Zhou, Jiajun Zhang, Shaonan Wang, and
  Chengqing Zong.
\newblock {NCLS}: Neural cross-lingual summarization.
\newblock In \emph{Proceedings of the 2019 Conference on Empirical Methods in
  Natural Language Processing and the 9th International Joint Conference on
  Natural Language Processing (EMNLP-IJCNLP)}, pages 3054--3064, Hong Kong,
  China, November 2019. Association for Computational Linguistics.
\newblock \doi{10.18653/v1/D19-1302}.
\newblock URL \url{https://aclanthology.org/D19-1302}.

\bibitem[Hermann et~al.(2015)Hermann, Kocisky, Grefenstette, Espeholt, Kay,
  Suleyman, and Blunsom]{hermann2015teaching}
Karl~Moritz Hermann, Tomas Kocisky, Edward Grefenstette, Lasse Espeholt, Will
  Kay, Mustafa Suleyman, and Phil Blunsom.
\newblock Teaching machines to read and comprehend.
\newblock \emph{Advances in neural information processing systems}, 28, 2015.

\bibitem[Zhu et~al.(2018)Zhu, Li, Liu, Zhou, Zhang, and Zong]{zhu2018msmo}
Junnan Zhu, Haoran Li, Tianshang Liu, Yu~Zhou, Jiajun Zhang, and Chengqing
  Zong.
\newblock {MSMO}: Multimodal summarization with multimodal output.
\newblock In \emph{Proceedings of the 2018 Conference on Empirical Methods in
  Natural Language Processing}, pages 4154--4164, Brussels, Belgium,
  October-November 2018. Association for Computational Linguistics.
\newblock \doi{10.18653/v1/D18-1448}.
\newblock URL \url{https://aclanthology.org/D18-1448}.

\bibitem[Auer(2005)]{auer2005postscript}
Peter Auer.
\newblock A postscript: Code-switching and social identity.
\newblock \emph{Journal of pragmatics}, 37\penalty0 (3):\penalty0 403--410,
  2005.

\bibitem[Hickey(2020)]{hickey2020handbook}
Raymond Hickey.
\newblock \emph{The handbook of language contact}.
\newblock John Wiley \& Sons, 2020.

\bibitem[Blevins and Zettlemoyer(2022)]{blevins2022language}
Terra Blevins and Luke Zettlemoyer.
\newblock Language contamination explains the cross-lingual capabilities of
  english pretrained models.
\newblock \emph{arXiv preprint arXiv:2204.08110}, 2022.

\bibitem[Blevins et~al.(2022)Blevins, Gonen, and
  Zettlemoyer]{blevins2022analyzing}
Terra Blevins, Hila Gonen, and Luke Zettlemoyer.
\newblock Analyzing the mono-and cross-lingual pretraining dynamics of
  multilingual language models.
\newblock \emph{arXiv preprint arXiv:2205.11758}, 2022.

\bibitem[Ouyang et~al.(2022)Ouyang, Wu, Jiang, Almeida, Wainwright, Mishkin,
  Zhang, Agarwal, Slama, Ray, et~al.]{ouyang2022training}
Long Ouyang, Jeff Wu, Xu~Jiang, Diogo Almeida, Carroll~L Wainwright, Pamela
  Mishkin, Chong Zhang, Sandhini Agarwal, Katarina Slama, Alex Ray, et~al.
\newblock Training language models to follow instructions with human feedback.
\newblock \emph{arXiv preprint arXiv:2203.02155}, 2022.

\bibitem[Shen et~al.(2019)Shen, Suzuki, Inui, Su, Klakow, and
  Sekine]{shen2019select}
Xiaoyu Shen, Jun Suzuki, Kentaro Inui, Hui Su, Dietrich Klakow, and Satoshi
  Sekine.
\newblock Select and attend: Towards controllable content selection in text
  generation.
\newblock In \emph{Proceedings of the 2019 Conference on Empirical Methods in
  Natural Language Processing and the 9th International Joint Conference on
  Natural Language Processing (EMNLP-IJCNLP)}, pages 579--590, 2019.

\bibitem[Tjoa and Guan(2020)]{tjoa2020survey}
Erico Tjoa and Cuntai Guan.
\newblock A survey on explainable artificial intelligence (xai): Toward medical
  xai.
\newblock \emph{IEEE transactions on neural networks and learning systems},
  32\penalty0 (11):\penalty0 4793--4813, 2020.

\bibitem[Burkart and Huber(2021)]{burkart2021survey}
Nadia Burkart and Marco~F Huber.
\newblock A survey on the explainability of supervised machine learning.
\newblock \emph{Journal of Artificial Intelligence Research}, 70:\penalty0
  245--317, 2021.

\bibitem[Narang et~al.(2020)Narang, Raffel, Lee, Roberts, Fiedel, and
  Malkan]{narang2020wt5}
Sharan Narang, Colin Raffel, Katherine Lee, Adam Roberts, Noah Fiedel, and
  Karishma Malkan.
\newblock Wt5?! training text-to-text models to explain their predictions.
\newblock \emph{arXiv preprint arXiv:2004.14546}, 2020.

\bibitem[Wiegreffe et~al.(2022)Wiegreffe, Hessel, Swayamdipta, Riedl, and
  Choi]{wiegreffe2022reframing}
Sarah Wiegreffe, Jack Hessel, Swabha Swayamdipta, Mark Riedl, and Yejin Choi.
\newblock Reframing human-{AI} collaboration for generating free-text
  explanations.
\newblock In \emph{Proceedings of the 2022 Conference of the North American
  Chapter of the Association for Computational Linguistics: Human Language
  Technologies}, pages 632--658, Seattle, United States, July 2022. Association
  for Computational Linguistics.
\newblock \doi{10.18653/v1/2022.naacl-main.47}.
\newblock URL \url{https://aclanthology.org/2022.naacl-main.47}.

\bibitem[Lampinen et~al.(2022)Lampinen, Dasgupta, Chan, Matthewson, Tessler,
  Creswell, McClelland, Wang, and Hill]{lampinen2022can}
Andrew~K Lampinen, Ishita Dasgupta, Stephanie~CY Chan, Kory Matthewson,
  Michael~Henry Tessler, Antonia Creswell, James~L McClelland, Jane~X Wang, and
  Felix Hill.
\newblock Can language models learn from explanations in context?
\newblock \emph{arXiv preprint arXiv:2204.02329}, 2022.

\bibitem[Kadavath et~al.(2022)Kadavath, Conerly, Askell, Henighan, Drain,
  Perez, Schiefer, Dodds, DasSarma, Tran-Johnson, et~al.]{kadavath2022language}
Saurav Kadavath, Tom Conerly, Amanda Askell, Tom Henighan, Dawn Drain, Ethan
  Perez, Nicholas Schiefer, Zac~Hatfield Dodds, Nova DasSarma, Eli
  Tran-Johnson, et~al.
\newblock Language models (mostly) know what they know.
\newblock \emph{arXiv preprint arXiv:2207.05221}, 2022.

\end{thebibliography}
\bibliographystyle{unsrtnat}

%%%%%%%%%%%%%%%%%%%%%%%%%%%%%%%%%%%%%%%%%%%%%%%%%%%%%%%%%%%%%%%%%%%%%%%%%%%%%%%
%%%%%%%%%%%%%%%%%%%%%%%%%%%%%%%%%%%%%%%%%%%%%%%%%%%%%%%%%%%%%%%%%%%%%%%%%%%%%%%
% APPENDIX
%%%%%%%%%%%%%%%%%%%%%%%%%%%%%%%%%%%%%%%%%%%%%%%%%%%%%%%%%%%%%%%%%%%%%%%%%%%%%%%
%%%%%%%%%%%%%%%%%%%%%%%%%%%%%%%%%%%%%%%%%%%%%%%%%%%%%%%%%%%%%%%%%%%%%%%%%%%%%%%

\end{document}